\documentclass[final,twocolumn,5p]{elsarticle}

\makeatletter
\def\ps@pprintTitle{%
 \let\@oddhead\@empty
 \let\@evenhead\@empty
 \def\@oddfoot{}%
 \let\@evenfoot\@oddfoot}
\makeatother

\usepackage{lineno,hyperref}
\modulolinenumbers[5]

\usepackage{import}
\usepackage{booktabs}
\usepackage{comment}
\usepackage{amssymb,amsmath}
\usepackage{graphicx}
\usepackage{color}
\usepackage[table]{xcolor}
\usepackage{makecell}
\usepackage{adjustbox}
\usepackage{array}

\hypersetup{
    colorlinks=true,
    linkcolor=blue,
}

\newcommand{\bftab}{\fontseries{b}\selectfont}
\newcommand{\cm}{\checkmark}
\newcommand{\myparagraph}[1]{\vspace{.2cm} \noindent \textbf{#1} \:}

\usepackage{etoolbox}
\makeatletter
\patchcmd{\subsection}{\itshape}{\bfseries}{}{}
\patchcmd{\subsubsection}{\itshape}{\bfseries}{}{}
\makeatother

\newcolumntype{R}[2]{%
    >{\adjustbox{angle=#1,lap=\width-(#2)}\bgroup}%
    c%
    <{\egroup}%
}
\newcommand*\rot{\multicolumn{1}{R{90}{1em}}}%

\begin{document}

\begin{frontmatter}

\title{Adversarial Text-to-Image Synthesis: A Review}

\author[1,2]{Stanislav Frolov\corref{correspondingauthor}}
\cortext[correspondingauthor]{Correspondence to: Deutsches Forschungszentrum f{\"u}r K{\"u}nstliche Intelligenz (DFKI), Research Department Smart Data \& Knowledge Services, Trippstadter Str. 122, 67663 Kaiserslautern, Germany}
\ead{stanislav.frolov@dfki.de}

\author[3]{Tobias Hinz}
\author[2]{Federico Raue}
\author[2]{J{\"o}rn Hees}
\author[1,2]{Andreas~Dengel}

\address[1]{Technische Universit{\"a}t Kaiserslautern, Germany}
\address[2]{Deutsches Forschungszentrum für K{\"u}nstliche Intelligenz (DFKI), Germany}
\address[3]{Universit{\"a}t Hamburg, Germany}

\begin{abstract}
With the advent of generative adversarial networks, synthesizing images from textual descriptions has recently become an active research area.
It is a flexible and intuitive way for conditional image generation with significant progress in the last years regarding visual realism, diversity, and semantic alignment.
However, the field still faces several challenges that require further research efforts such as enabling the generation of high-resolution images with multiple objects, and developing suitable and reliable evaluation metrics that correlate with human judgement.
In this review, we contextualize the state of the art of adversarial text-to-image synthesis models, their development since their inception five years ago, and propose a taxonomy based on the level of supervision.
We critically examine current strategies to evaluate text-to-image synthesis models, highlight shortcomings, and identify new areas of research, ranging from the development of better datasets and evaluation metrics to possible improvements in architectural design and model training.
This review complements previous surveys on generative adversarial networks with a focus on text-to-image synthesis which we believe will help researchers to further advance the field.
\end{abstract}


\end{frontmatter}

\section{Introduction}

When humans hear or read a story, they immediately draw mental pictures visualizing the content in their head.
The ability to visualize and understand the intricate relationship between the visual world and language is so natural that we rarely think about it.
Visual mental imagery or ``seeing with the mind’s eye'' also plays an important role in many cognitive processes such as memory, spatial navigation, and reasoning \cite{Kosslyn2001NeuralFO}.
Inspired by how humans visualize scenes, building a system that understands the relationship between vision and language, and that can create images reflecting the meaning of textual descriptions, is a major milestone towards human-like intelligence.

In the last few years, computer vision applications and image processing techniques have greatly benefited from advancements enabled by the breakthrough of deep learning.
One of these is the field of image synthesis which is the process of generating new images and manipulating existing ones.
Image synthesis is an interesting and important task because of many practical applications such as art generation, image editing, virtual reality, video games, and computer-aided design.

The advent of Generative Adversarial Networks (GANs) \cite{GoodfellowGAN2014} made it possible to train generative models for images in a completely unsupervised manner.
GANs have sparked a lot of interest and advanced research efforts in synthesising images.
They framed the image synthesis task as a two-player game of two competing artificial neural networks.
A generator network is trained to produce realistic samples, while a discriminator network is trained to distinguish between real and generated images.
The training objective of the generator is to fool the discriminator.
This approach has successfully been adapted to many applications such as high-resolution synthesis of human faces \cite{Karras2018ProgressiveGO}, image super-resolution \cite{super-resolution}, image in-painting \cite{image-inpainting,yu2019free}, data augmentation \cite{frid2018gan}, style transfer \cite{Gatys2016ImageST,Jing2019NeuralST}, image-to-image translation \cite{Isola_2017_CVPR,Zhu_2017_ICCV}, and representation learning \cite{Bengio2013RepresentationLA,Donahue2019LargeSA}.

Further developments in this field allowed the extension of these approaches to learn conditional generative models \cite{cGANMirza2014}.
Motivated by how humans draw mental pictures, an intuitive interface for conditional image synthesis can be achieved by using textual descriptions.
Compared to labels, textual descriptions can carry dense semantic information about the present objects, their attributes, spatial arrangements, relationships, and allow to represent diverse and detailed scenes.

In this review, we focus on text-to-image (T2I) synthesis, which aims to produce an image that correctly reflects the meaning of a textual description.
T2I can be seen as the inverse of image captioning \cite{Hossain2019ACS}, where the input is an image and the output is a textual description of that image.
Although the methods presented in this review can be applied to many image domains, most T2I research focuses on methods generating visually realistic, photographic, natural images.

The field of purely generative text-to-image synthesis was started by the work of Reed et al. in 2016 \cite{Reed2016}.
It extended conditional GANs to generate natural images based on textual descriptions and was shown to work on restricted datasets (e.g., Oxford-102 Flowers \cite{Nilsback2008} and CUB-200 Birds \cite{WahCUB_200_2011}) and relatively small image resolution ($64 \times 64$ pixels).
In the last five years, this field has seen large improvements both in the quality of generated images (based on qualitative and quantitative evaluation), the complexity of used data sets (e.g., COCO \cite{COCO}), and the resolution of generated images (e.g., $256 \times 256$ and higher).
Some of these advancements include approaches such as improved text encodings, loss-terms introduced specifically for text-to-image synthesis, and novel architectures (e.g., stacked networks and attention).
Furthermore, the field has developed quantitative evaluation metrics (e.g., R-precision, Visual-Semantic similarity, and Semantic Object Accuracy) that were introduced specifically to evaluate the quality of text-to-image synthesis models.

However, the field still faces several challenges.
Despite much progress, current models are still far from being capable of generating complex scenes with multiple objects based only on textual descriptions.
There is also very limited work on scaling these approaches to resolutions higher than $256 \times 256$ pixels.
We also find that it is challenging to reproduce the quantitative results of many approaches, even if code and pre-trained models are provided.
This is reflected in the literature where often different quantitative results are reported for the same model.
Furthermore, we observe that many of the currently used evaluation metrics are unsuitable for evaluating text-to-image synthesis models and do not correlate well with human perception.
This is amplified by the fact that only a few approaches perform human user studies to assess if their improvements are evident in a qualitative sense, and if they do, the studies are not standardized, making the comparison of results difficult.

This review aims to highlight and contextualize the development of the current state of the art of generative T2I models and their development since their inception five years ago.
We give an outline of where the research is currently headed and where more work is needed from the community.
We critically examine the current approach to evaluating T2I models and highlight several shortcomings in current metrics.
Finally, we identify new areas of research for T2I models, ranging from the development of better datasets and evaluation metrics to possible improvements in architectural design and model training.
In contrast to previous surveys and reviews \cite{WuSurvey,GansOverview,HongSurvey,Wang2020ASR,mogadala2019trends}, this review specifically focuses on the development and evaluation of T2I methods.
This review also goes beyond the only other existing T2I survey \cite{T2ISurveyTaxonomy} by incorporating more approaches, thoroughly discussing the current state of evaluation techniques in the T2I field and systematically examining open challenges.

\begin{figure}[t]
    \begin{center}
    \includegraphics[scale=0.4]{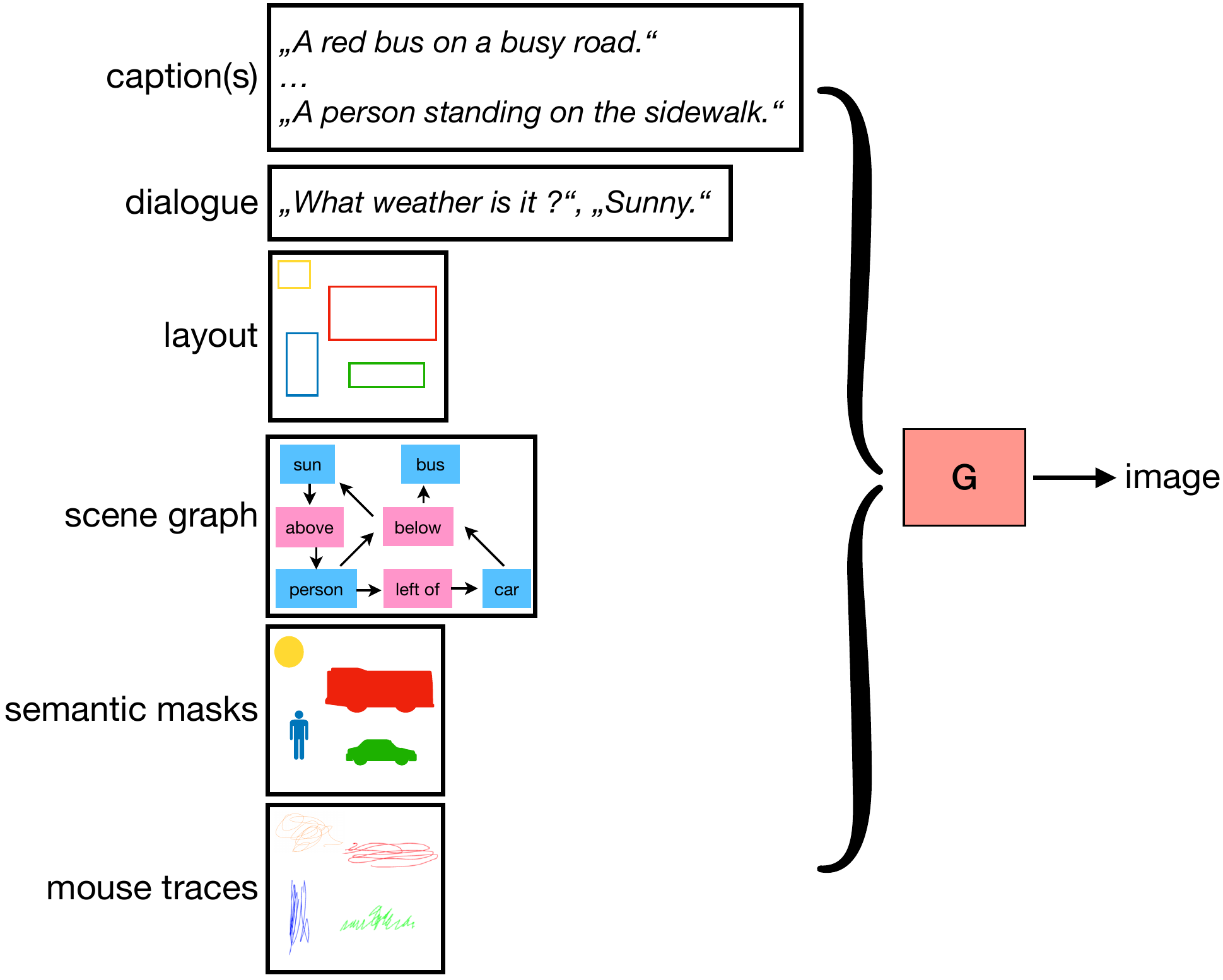}
    \caption{
    Overview of annotations that have been used to generate images from text.
    We first revisit direct T2I methods which use single captions as input.
    Next, we discuss approaches that leverage additional information as input.
    }
    \label{fig:additional_inputs_methods}
    \end{center}
\end{figure}

We first revisit the fundamentals of GANs, commonly used datasets and text encoders that produce the embedding of a textual description for conditioning (\autoref{fundamentals}).
After this, we propose a taxonomy of methods based on the level of supervision during training, namely approaches for direct T2I synthesis that use single captions as input (\autoref{main-section}) versus approaches that use additional information such as multiple captions, dialogue, layout, scene graphs, or masks (\autoref{additional_supervision}).
See \autoref{fig:additional_inputs_methods} for an overview of annotations that haven been used for T2I synthesis.
Next, we specifically focus on evaluation techniques used by the T2I community and revisit image-quality and image-text alignment metrics as well as how user studies are conducted (\autoref{evaluation}).
We gather published results, highlight and identify challenges associated with using these evaluation strategies, define desiderata for future metrics and suggest how to use currently available metrics to assess the performance of T2I models.
Finally, we offer a thorough discussion of the state of the art across multiple dimensions such as the suitability of datasets, choice and developments of model architectures, evaluation metrics, and on-going as well as possible future research directions (\autoref{discussion}).
Complementing other reviews on generative models, we believe that our work will help tackle open challenges and further advance the field.

\section{Fundamentals}\label{fundamentals}
This section revisits four key components required to understand the T2I methods discussed in the next sections: the original (unconditional) GAN \cite{GoodfellowGAN2014} that takes noise as input to produce an image, the conditional GAN (cGAN) \cite{cGANMirza2014} which allows to condition the generated image on a label, text encoders used to produce the embedding of a textual description for conditioning, and commonly used datasets by the T2I community.

\subsection{Generative Adversarial Networks}
\begin{figure}[t]
    \begin{center}
    \includegraphics[scale=0.3]{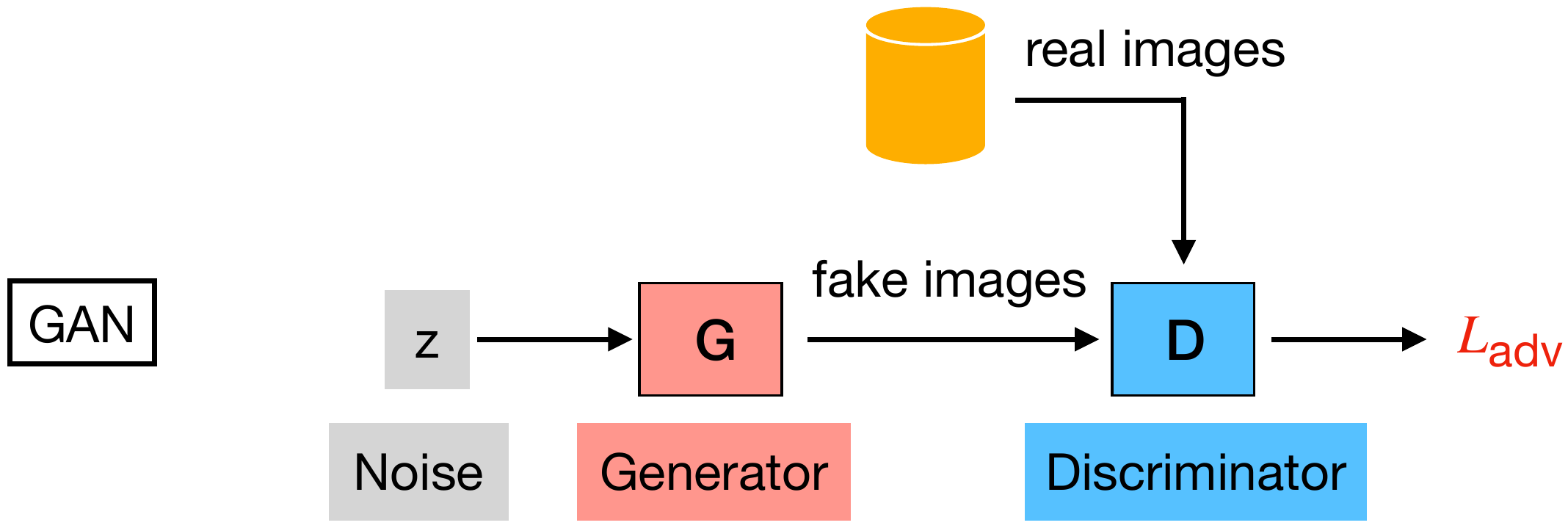}
    \caption{
    Simplified architecture of a GAN \cite{GoodfellowGAN2014}.
    Given noise input $z$ randomly sampled from a normal distribution, the generator is trained to produce images to fool the discriminator.
    The discriminator is trained to distinguish between real and generated images.
    }
    \label{fig:gan}
    \end{center}
\end{figure}

The original GAN proposed in \cite{GoodfellowGAN2014} consists of two neural networks: a generator network $G(z)$ with noise $z\sim p_z$ sampled from a prior noise distribution, and a discriminator network $D(x)$, where $x\sim p_{data}$ are real, and $x\sim p_{g}$ are generated images, respectively.
Training is formulated as a two-player game in which the discriminator is trained to distinguish between real and generated images, while the generator is trained to capture the real data distribution and produce images to fool the discriminator.
See \autoref{fig:gan} for an illustration of the GAN architecture.

More formally, as in \cite{GoodfellowGAN2014}, the training can be defined as a two-player minimax game with the value function $V(D,G)$, where the discriminator $D(x)$ is trained to maximize the log-likelihood it assigns to the correct class, while the generator $G(z)$ is trained to minimize the probability being classified as fake by the discriminator $\text{log}(1-D(G(z))$, see \autoref{eq:gan}. The loss function is indicated as $L_\text{adv}$ in our figures.
\begin{align}
\begin{aligned}\label{eq:gan}
  \underset{G}{\min}\,\underset{D}{\max}\,V(D,G) &=\mathbb{E}_{x\sim p_{data}}[\log D(x)] \\
  & + \mathbb{E}_{z\sim p_z}[\log(1-D(G(z))]
  \end{aligned}
\end{align}

\subsection{Conditional GANs}
\begin{figure}[t]
    \begin{center}
    \includegraphics[scale=0.35]{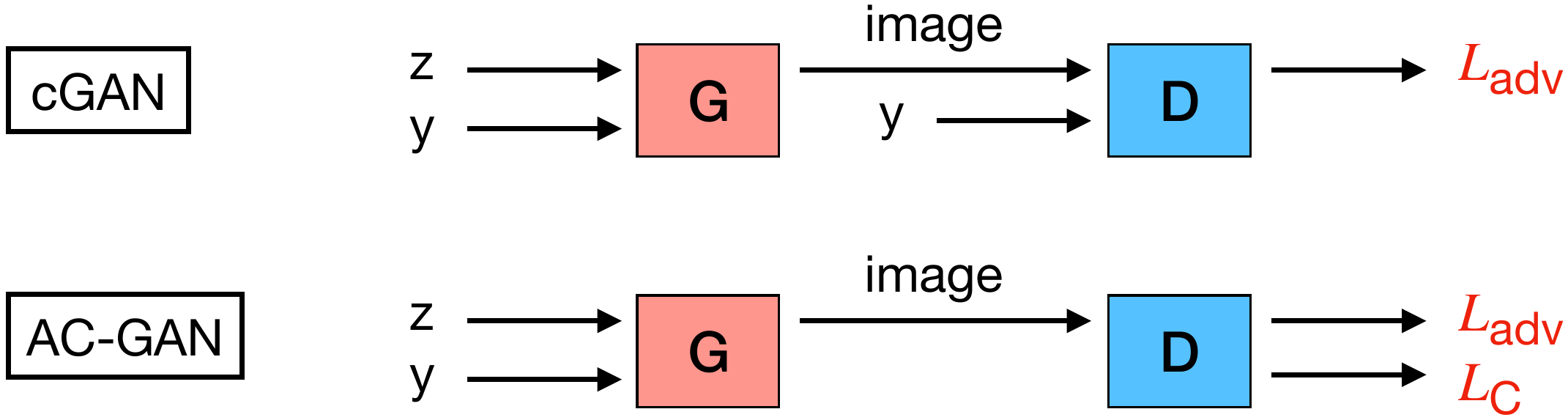}
    \caption{
    Simplified cGAN \cite{cGANMirza2014} and AC-GAN \cite{auxOdena} architectures.
    In cGAN \cite{cGANMirza2014}, the class label is input to both generator and discriminator networks.
    In AC-GAN \cite{auxOdena}, the discriminator is trained with an additional auxiliary classification loss.
    Note that we are omitting to depict real images as input to the discriminator in the following figures for brevity.
    }
    \label{fig:conditional_gans}
    \end{center}
\end{figure}

Although generating new, realistic samples is interesting, gaining control over the image generation process has high practical value.
Mirza et al. proposed the conditional GAN (cGAN) \cite{cGANMirza2014} by incorporating a conditioning variable $y$ (e.g., class labels) at both the generator and discriminator to specify which MNIST \cite{mnist} digit to produce.
See \autoref{fig:conditional_gans} for an illustration.
In their experiments, $z \sim p_z$ and $y$ are inputs to a Multi-Layer Perceptron (MLP) network with one hidden layer, thereby forming a joint hidden representation for the generator.
Analogously, for the discriminator, an MLP combines images and labels.
As given in \cite{cGANMirza2014}, \autoref{eq:gan} becomes \autoref{eq:cgan}.

\begin{align}
\begin{aligned}\label{eq:cgan}
  \underset{G}{\min}\,\underset{D}{\max}\,V(D,G) &= \mathbb{E}_{x\sim p_{data}}[\log D(x|y)] \\
  &+\mathbb{E}_{z\sim p_z}[\log (1-D(G(z|y))]
\end{aligned}
\end{align}

A number of variants extended the cGAN objective function to improve conditional GAN training.
For example, in AC-GAN \cite{auxOdena} the authors proposed adding an auxiliary classification loss to the discriminator, indicated as $L_\text{C}$ in \autoref{fig:conditional_gans}.

\subsection{Encoding Text}\label{encoding_text}
Creating an embedding from textual representations that is useful for the network in terms of a conditioning variable is not trivial.
Reed et al. \cite{Reed2016a} obtain the text encoding of a textual description by using a pre-trained character-level convolutional recurrent neural network (char-CNN-RNN).
The char-CNN-RNN is pre-trained to learn a correspondence function between text and image based on the class labels.
This leads to visually discriminative text encodings.
During training, additional text embeddings were generated by simply interpolating between the embeddings of two training captions.
The authors also showed that traditional text representations such as Word2Vec \cite{Word2Vec} and Bag-of-Words \cite{BoW} were less effective.
TAC-GAN \cite{Dash2017} employed Skip-Thought vectors \cite{Kiros2015SkipThoughtV}.

Instead of using the fixed text embedding obtained by a pre-trained text encoder, the authors of StackGAN \cite{Zhang2017} proposed Conditioning Augmentation (CA) to randomly sample the latent variable from a Gaussian distribution where the mean and covariance matrix are functions of the text embedding.
The Kullback-Leibler (KL) divergence between a standard Gaussian distribution and the conditioning Gaussian distribution is used as a regularization term during training.
This technique yields more training pairs and encourages smoothness over the conditioning manifold.
Many of the following T2I methods adopted this technique.
Similar to CA, in \cite{Souza2020EfficientNA} the authors proposed Sentence Interpolation (SI), a deterministic way to provide a continuous and smooth embedding space during training.

\begin{table*}[t]
\small
\begin{center}
    \begin{tabular}{lrrrrr}
        \toprule
        Dataset                                & \makecell[r]{Training \\ Images} & \makecell[r]{Testing \\ Images} & \makecell[r]{Total \\ Images} & \makecell[r]{Captions \\ per Image} & \makecell[r]{Object \\ Categories} \\
        \midrule
        Oxford-102 Flowers \cite{Nilsback2008}  & 7,034             & 1,155           & 8,189          & 10                & 102 \\
        CUB-200 Birds \cite{WahCUB_200_2011}    & 8,855             & 2,933           & 11,788         & 10                & 200 \\
        COCO \cite{COCO}                        & 82,783            & 40,504          & 123,287        & 5                 & 80 \\
        \bottomrule
    \end{tabular}
    \caption{
    Overview of commonly used datasets for T2I synthesis.
    }
    \label{table:datasets}
\end{center}
\end{table*}

The authors of AttnGAN \cite{Xu2018} replaced the char-CNN-RNN with a bi-directional LSTM (BiLSTM) \cite{schuster1997bidirectional} to extract feature vectors by concatenating the hidden states of the BiLSTM to form a feature matrix for each word.
The global sentence vector is formed by concatenating the last hidden states.
The text encoder is obtained by pre-training a Deep Attentional Multimodal Similarity Model (DAMSM) to compute word features that match image subregions (image-text similarity at the word level).
The BiLSTM is trained to match the intermediate features of a pre-trained image classifier.
Since the introduction of using BiLSTM in AttnGAN \cite{Xu2018} to encode captions, most of the following works adopted it.
However, recent works \cite{wang2020faces,Pavllo2020ControllingSA} leverage pre-trained transformer-based models such as BERT \cite{Devlin2019BERTPO} to obtain text embeddings.

\subsection{Datasets}

\begin{figure}[t]
    \begin{center}
    \includegraphics[scale=0.55]{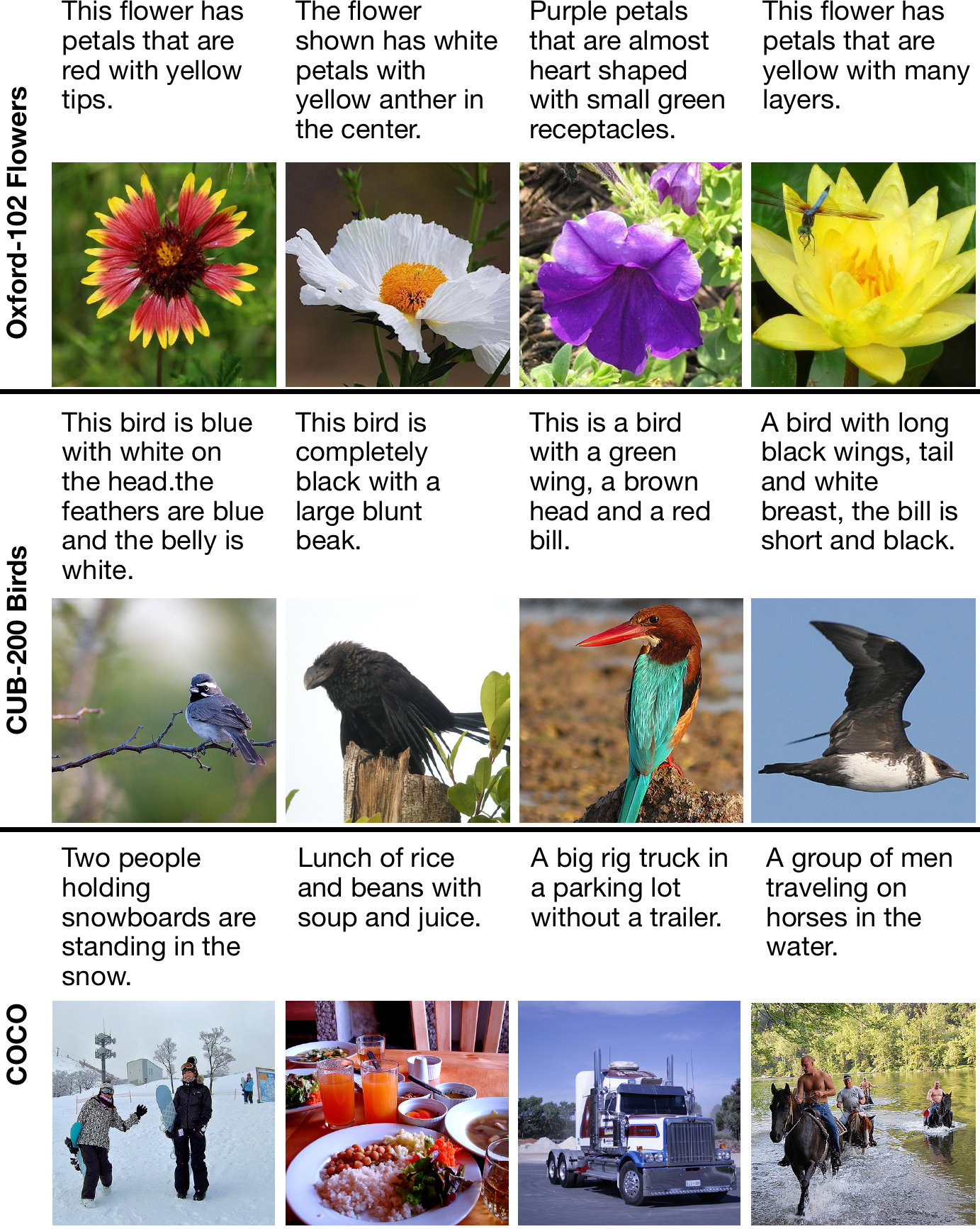}
    \caption{
    Example images and corresponding captions of common T2I datasets.
    }
    \label{fig:datset_examples}
    \end{center}
\end{figure}

Datasets are at the core of every machine learning problem.
Widely adopted datasets in T2I research are Oxford-120 Flowers \cite{Nilsback2008}, CUB-200 Birds \cite{WahCUB_200_2011}, and COCO \cite{COCO}.
Both Oxford-102 Flowers \cite{Nilsback2008} and CUB-200 Birds \cite{WahCUB_200_2011} are relatively small datasets containing around 10k images.
Each image depicts a single object and there are ten associated captions per image.
COCO \cite{COCO} on the other hand consists of around 123k images with five captions per image.
In contrast to both Oxford-102 Flowers and CUB-200 Birds, images in the COCO dataset usually contain multiple, often interacting objects in complex settings.
\autoref{table:datasets} shows an overview of the dataset statistics.
Most T2I works use the official 2014 COCO split.
Example images and corresponding captions are provided in \autoref{fig:datset_examples}.

\section{Direct T2I Methods}\label{main-section}
After revisiting GANs, text encoders and commonly used datasets in the previous chapter, we now review state-of-the-art methods for direct T2I.
We start with the first T2I approach by Reed at al. proposed in 2016, followed by the use of stacked architectures.
Next, we discuss the introduction of attention mechanisms, the use of  architectures, cycle consistency approaches, and the use of dynamic memory networks.
Finally, we discuss approaches that adapt unconditional models for T2I.

\subsection{First T2I Approaches}\label{initial_t2i}
\begin{figure}[t]
    \begin{center}
    \includegraphics[scale=0.3]{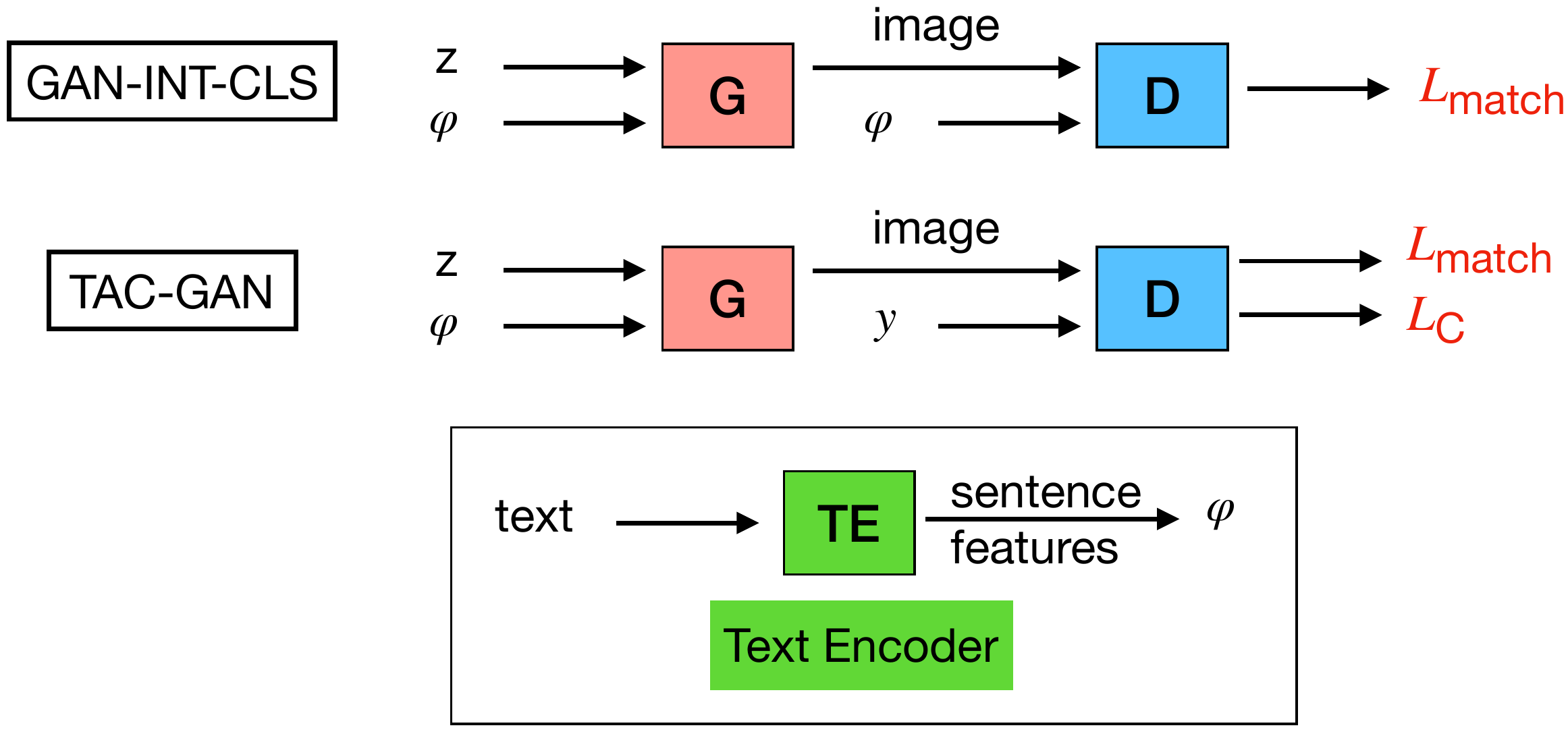}
    \caption{GAN-INT-CLS \cite{Reed2016} conditions both generator and discriminator on a text embedding provided by the pre-trained char-CNN-RNN text encoder and employs the matching aware pair loss $L_\text{match}$.
    TAC-GAN \cite{Dash2017} uses an additional auxiliary classification task and loss $L_\text{C}$ during training.
    }
    \label{fig:initial_t2i_gans}
    \end{center}
\end{figure}

The first T2I approach by Reed et al. \cite{Reed2016} conditions the generation process on the whole sentence embedding obtained from a pre-trained text encoder.
The discriminator is trained to distinguish between real and generated image-text pairs.
Hence, the first T2I model is a natural extension of a cGAN \cite{cGANMirza2014} in that the conditioning on a class label $y$ is simply replaced by a text embedding $\varphi$.
In GAN-INT-CLS \cite{Reed2016}, three different pairs are used as input to the discriminator: a real image with matching text, a generated image with corresponding text, and a real image with mismatching text.
This approach is often referred to as the matching aware discriminator and the corresponding objective is indicated as $L_\text{match}$ in our figures.
This approach forces both the generator and the discriminator to not only focus on realistic images but also to align them with the input text.
See \autoref{fig:initial_t2i_gans} for a simplified architecture.
Compared to GAN-INT-CLS \cite{Reed2016}, TAC-GAN \cite{Dash2017} employs an additional auxiliary classification loss inspired by AC-GAN \cite{auxOdena} using one-hot encoded class labels.

\subsection{Stacked Architectures}\label{stacked}

GAN-INT-CLS \cite{Reed2016} was able to generate low-resolution $64 \times 64$ pixel images, while TAC-GAN \cite{Dash2017} generated $128 \times 128$ pixel images.
In order to enable T2I models to synthesize higher resolution images, many following works proposed to use multiple, stacked generators.

In StackGAN \cite{Zhang2017}, the first stage generates a coarse $64 \times 64$ pixel image given a random noise vector and textual conditioning vector.
This initial image and the text embedding are then input to a second generator that outputs a $256 \times 256$ pixel image.
At both stages, a discriminator is trained to distinguish between matching and non-matching image-text pairs.

\begin{figure}[t]
    \begin{center}
    \includegraphics[scale=0.3]{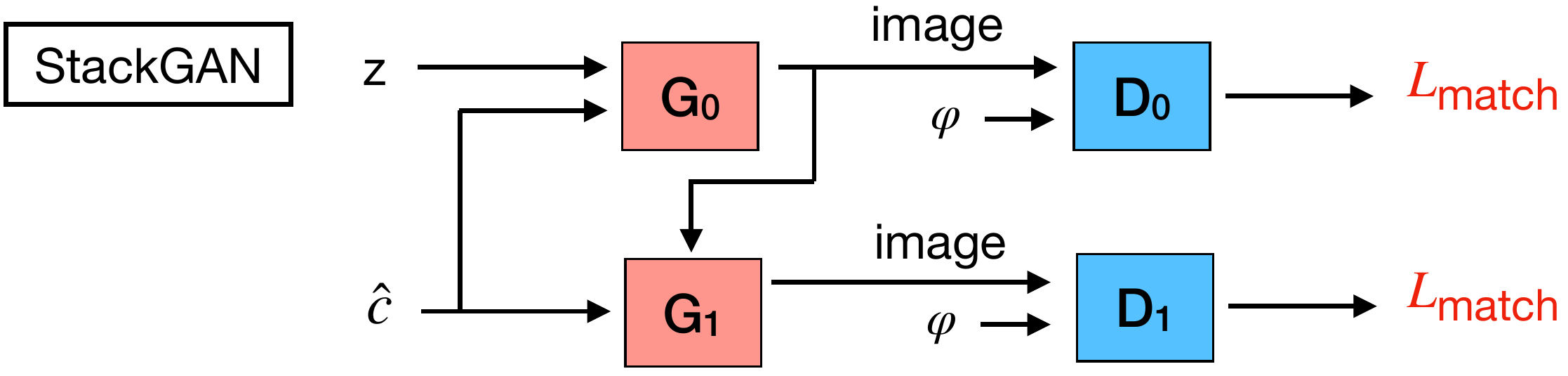}
    \includegraphics[scale=0.3]{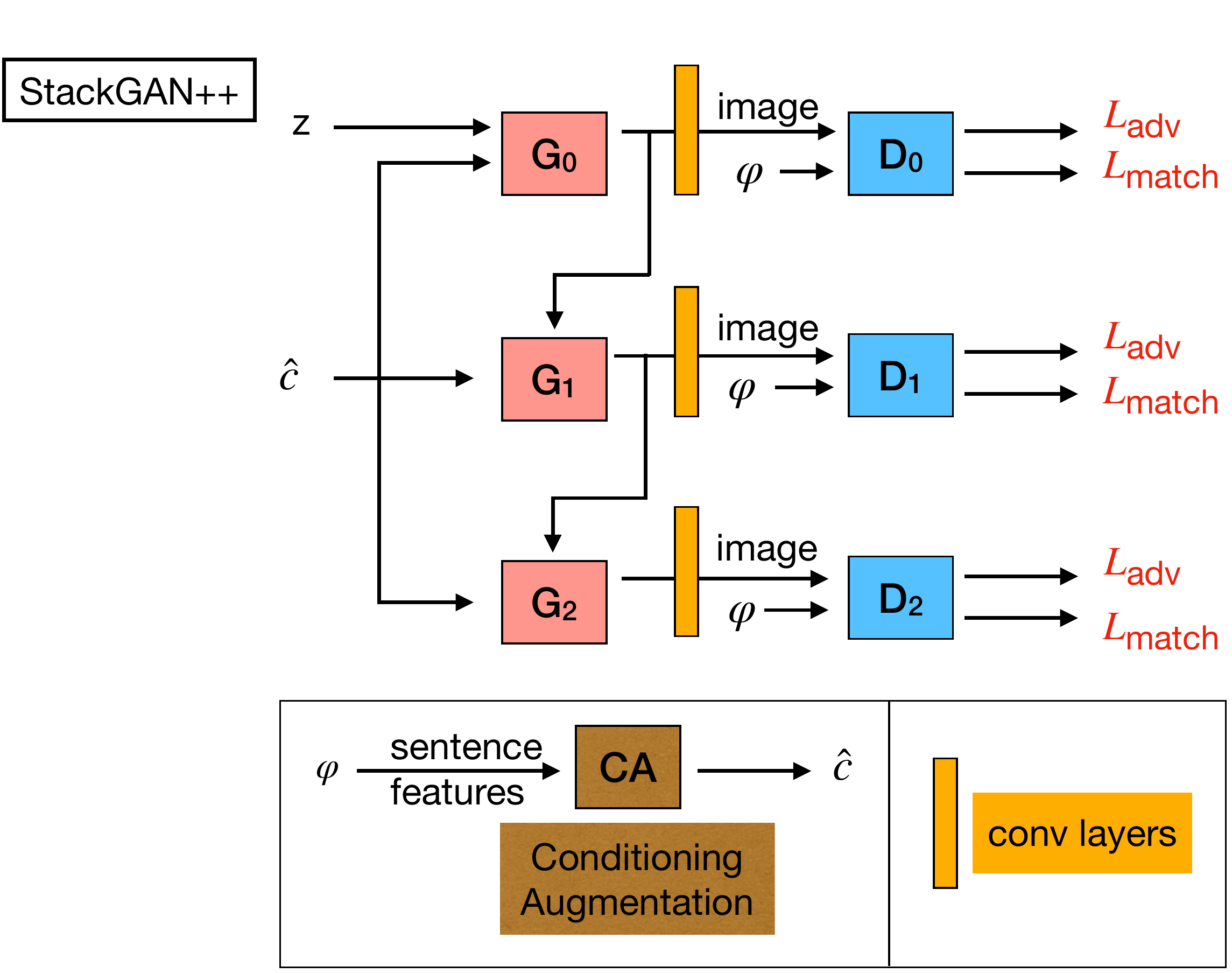}
    \caption{
    StackGAN \cite{Zhang2017} and StackGAN++ \cite{Zhang2019b} architectures.
    While StackGAN requires a two-stage training pipeline, StackGAN++ can be trained end-to-end.
    During training, intermediate visual features are passed as input to the next generator stage, while additional convolutional layers produce the image.
    CA: text embeddings $\hat{c}$ are sampled from a Gaussian distribution to provide a smooth conditioning manifold.
    }
    \label{fig:stackgans}
    \end{center}
\end{figure}

StackGAN++ \cite{Zhang2019b} improved the architecture further via an end-to-end framework in which three generators and discriminators are jointly trained to simultaneously approximate the multi-scale, conditional and unconditional image distributions.
The authors proposed to sample text embeddings from a Gaussian distribution for a smooth conditioning manifold, instead of using fixed text embeddings.
To encourage the network to produce images at each scale to share basic structure and colors, an additional color-consistency regularization term was proposed that aims at minimizing the differences between the mean and covariance of pixels between different scales.
\autoref{fig:stackgans} shows the architecture of StackGAN \cite{Zhang2017} and StackGAN++ \cite{Zhang2019b}.

Similar to the idea of training conditional and unconditional distributions at the same time, FusedGAN \cite{Bodla2018a} consists of two generators (one for unconditional and one for conditional image synthesis) that partly share a common latent space to allow both conditional and unconditional generation from the same generator.

To overcome the need for multiple generator networks, HDGAN \cite{Zhang2018d} employed hierarchically-nested discriminators at multi-scale intermediate layers to generate $512 \times 512$ images.
In other words, the adversarial game is played along the depth of the generator with distinct discriminators at each level of resolution.
In addition to the matching aware pair loss, the discriminators are also trained to distinguish real from generated image patches.
This objective acts as a regularizer to the hidden layers of the generator, since outputs at intermediate layers can utilize the signal from discriminators at higher resolutions to produce more consistent outputs between different scales.

Similarly, PPAN \cite{Gao2019} uses only one generator and three distinct discriminators.
The generator of PPAN applies a pyramid framework \cite{Lin2017FeaturePN,Lai2017DeepLP} to combine low-resolution, semantically strong features with high-resolution, semantically weak features through a down-to-top pathway with lateral connections.
During training the authors additionally employed a perceptual loss \cite{Ledig2017PhotoRealisticSI} based on extracted features from a pre-trained VGG \cite{Simonyan2015VeryDC} network, and an auxiliary classification loss.

In contrast, HfGAN \cite{HfGAN} uses a hierarchically-fused architecture with only one discriminator.
Multi-scale global features are extracted from different stages and adaptively fused together such that lower-resolution feature maps that are spatially coarse, but contain and dictate the overall semantic structure of the generated image, can guide the generation of fine details.
Inspired by ResNet \cite{He2016DeepRL}, the authors adopted identity addition, weighted addition, and shortcut connections as their fusion method.

\subsection{Attention Mechanisms}\label{attention}

\begin{figure}[t]
    \begin{center}
    \includegraphics[scale=0.3]{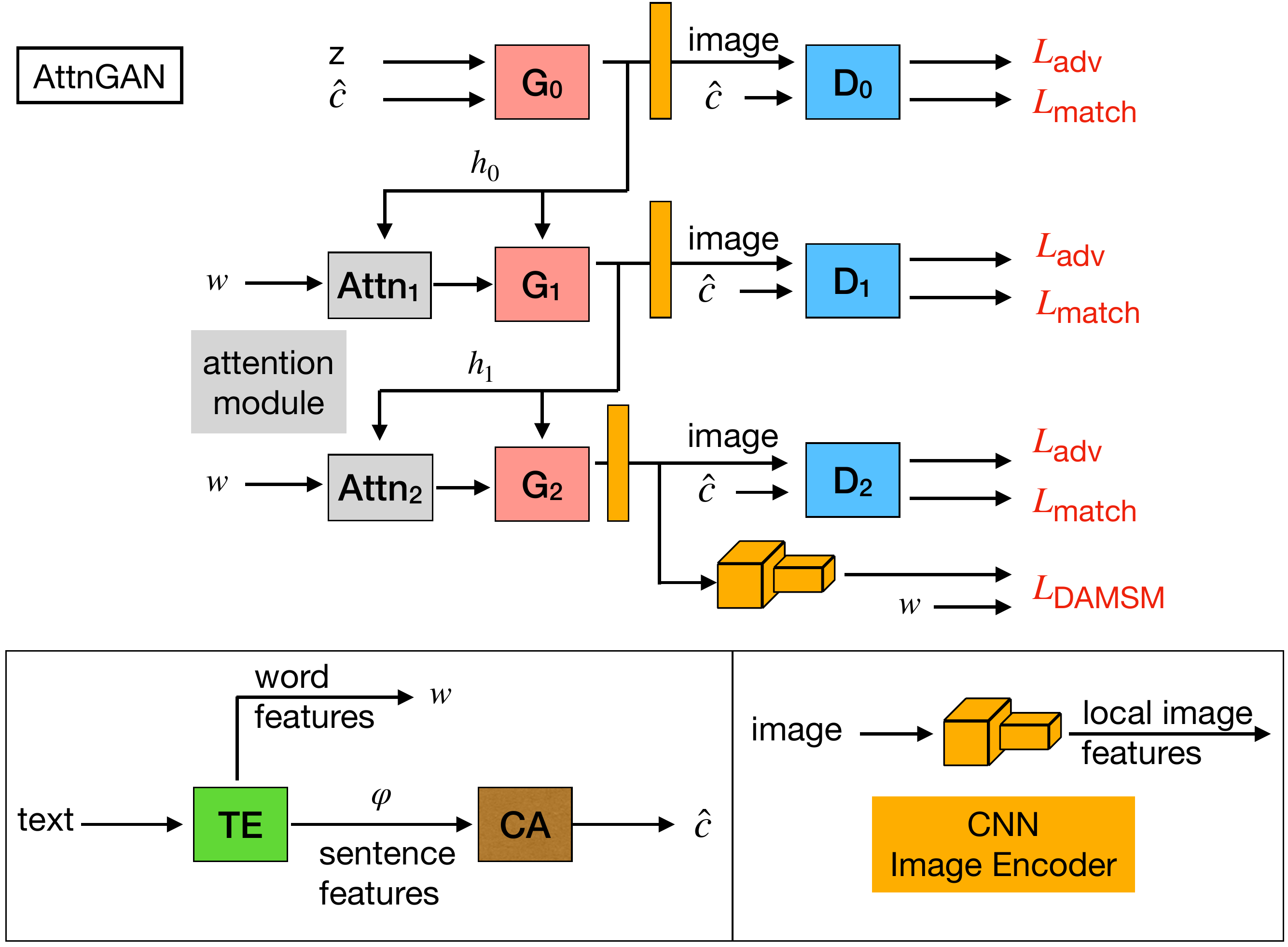}
    \caption{
    Simplified AttnGAN \cite{Xu2018} architecture.
    The attention modules and similarity loss between local image and word features $L_\text{DAMSM}$ help the generator to synthesize fine-grained details based on relevant words.
    }
    \label{fig:attngan}
    \end{center}
\end{figure}

Attention techniques allow the network to focus on specific aspects of an input by weighting important parts more than unimportant parts.
Attention is a very powerful technique and had a major impact on improving language and vision applications \cite{Bahdanau2015NeuralMT,Xu2015ShowAA,Luong2015EffectiveAT,Vaswani2017AttentionIA}.
AttnGAN \cite{Xu2018} builds upon StackGAN++ \cite{Zhang2019b} and incorporates attention into a multi-stage refinement pipeline.
The attention mechanism allows the network to synthesize fine-grained details based on relevant words in addition to the global sentence vector.
During generation, the network is encouraged to focus on the most relevant words for each sub-region of the image.
This is achieved via the Deep Attentional Multimodal Similarity Model (DAMSM) loss during training that computes the similarity between generated image and input text using both sentence and word level information.
See \autoref{fig:attngan} for an illustration of AttnGAN.

Huang et al. \cite{Huang2019RealisticIG} extended grid-based attention with an additional mechanism between object-grid regions and word phrases, where the object-grid regions are defined by auxiliary bounding boxes.
Phrase features are extracted in addition to sentence and word features by applying part-of-speech tagging.

The authors of SEGAN \cite{Tan2019b} proposed an attention competition module to focus only on key-words instead of defining an attention weight for each word in the sentence (as is done in AttnGAN).
They achieved this by introducing an attention regularization term (inspired by \cite{Lin2017ASS,Li2018DiversityRS}) that only keeps the attention weights for visually important words.

ControlGAN \cite{Li2019f} can do both: T2I generation and manipulation of visual attributes such as category, texture, and colour by changing the description without affecting other content (e.g., background and pose).
The authors proposed a word-level spatial and channel-wise attention-driven generator which allows the generator to synthesize image regions corresponding to the most relevant words.
Compared to the spatial attention in \cite{Xu2018} which mainly focuses on colour information, the channel-wise attention correlates semantically meaningful parts with corresponding words (e.g., ``head'' and ``wings'' for CUB-200 birds).
A word-level discriminator provides the generator with fine-grained training signals and disentangles different visual attributes by exploiting the correlation between words and image subregions.

\subsection{Siamese Architectures}\label{siamese}
Siamese networks, first proposed to solve signature \cite{Siamese} and face verification problems \cite{Chopra2005LearningAS}, typically consist of two branches with shared model parameters operating on a pair of inputs.
Each branch operates on a different input, and the goal is to attain a mapping where inputs with similar patterns are placed more closely to each other than dissimilar ones.

\begin{figure}[t]
    \begin{center}
    \includegraphics[scale=0.3]{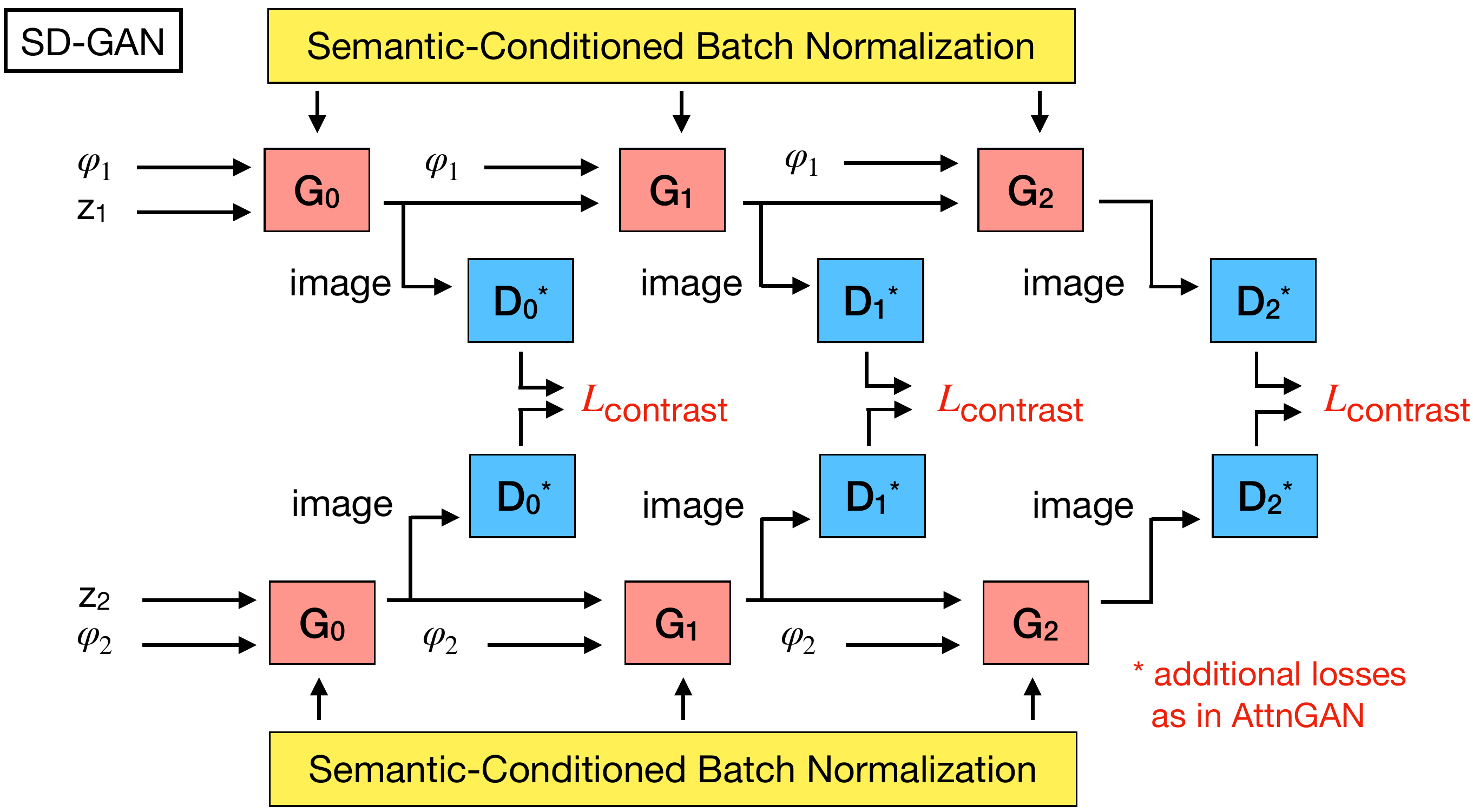}
    \caption{
    Simplified SD-GAN \cite{Yin2019} architecture.
    Depending on whether the two captions input to each of the branches are from the same ground truth image or not, the contrastive loss minimizes or maximizes the distance between the computed features to learn semantic commons.
    The Semantic-Conditioned Batch Normalization, a variant of conditional batch normalization \cite{Dumoulin2017ALR}, takes linguistic cues as input and adapts the visual feature maps.
    }
    \label{fig:sdgan}
    \end{center}
\end{figure}

SD-GAN \cite{Yin2019} is such a Siamese network architecture consisting of two branches.
While the individual branches of the network process different text inputs to produce an image, the model parameters are shared.
A contrastive loss based on \cite{ConstrastiveLoss} is employed to minimize / maximize the distance between the features computed in each branch to learn a semantically meaningful representation, depending on whether the two captions are from the same ground truth image (intra-class pair) or not (inter-class pair).
This approach distills semantic commons from text but might tend to ignore fine-grained semantic diversity.
In order to maintain the diversity in generated images, the authors additionally proposed Semantic-Conditioned Batch Normalization, a variant of conditional batch normalization \cite{Dumoulin2017ALR}, to adapt the visual feature maps depending on the linguistic cues.
See \autoref{fig:sdgan} for an illustration of SD-GAN.

SEGAN \cite{Tan2019b} trains a Siamese architecture to exploit ground truth images for semantic alignment.
They do so by minimizing the feature distance between generated image and corresponding ground truth image while maximizing the distance to another real image associated with a different caption.
To effectively balance easy versus hard samples, the authors proposed a sliding loss inspired by the focal loss \cite{Lin2020FocalLF} to adapt the relative importance of easy and hard sample pairs.

Instead of randomly sampling a mismatching negative image sample, in Text-SeGAN \cite{Cha2019} several strategies based on curriculum learning \cite{Bengio2009CurriculumL} are introduced to select negative samples with gradually increasing semantic difficulty.
Instead of using classification as an auxiliary task, the authors formulated a regression task to estimate semantic correctness based on the semantic distance to the encoded reference text.

\subsection{Cycle Consistency}\label{cycle}
We group T2I models that take the generated image and pass it through an image captioning \cite{PPGN,Qiao2019,Chen2019d} or image encoder network \cite{Lao2019}, thereby creating a cycle to the input description or latent code, as cycle consistency approaches.

PPGN \cite{PPGN} is based on feedback from a conditional network, which can either be a classifier or an image captioning network for conditional image synthesis.
The main idea is to iteratively find the latent code that leads the generator to produce an image which maximizes a specific feature activation in the feedback network (e.g., classification score or hidden vector of an RNN).
In this framework, a pre-trained generator can be re-purposed by plugging in a different feedback network.

\begin{figure}[t]
    \begin{center}
    \includegraphics[scale=0.3]{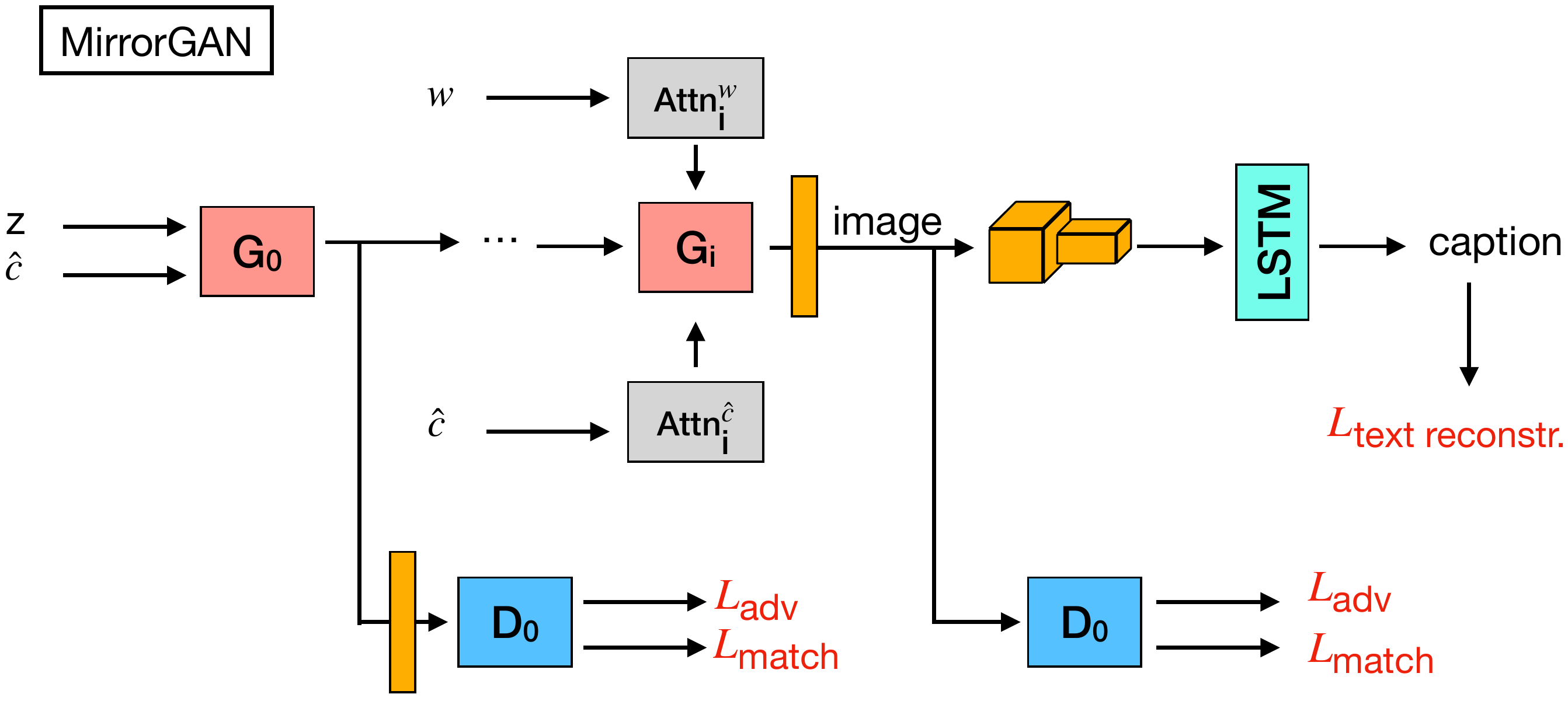}
    \caption{
    Simplified MirrorGAN \cite{Qiao2019} architecture.
    An image captioning network takes the generated image as input and produces a caption.
    A cross-entropy based text reconstruction loss aligns the generated caption to the input caption that was used to generate the image, thereby creating a cycle.
    }
    \label{fig:mirrogan}
    \end{center}
\end{figure}

Inspired by CycleGAN \cite{Zhu_2017_ICCV}, cycle-consistent image generation by re-description architectures \cite{Qiao2019,Chen2019d} learn a semantically consistent representation between text and image by appending a captioning network and train the network to produce a semantically similar caption from the synthesized image.
In MirrorGAN \cite{Qiao2019}, sentence and word embeddings are used to guide a cascaded generator architecture via both global sentence and local word attention.
Next, an encoder-decoder based image captioning network \cite{Karpathy2015DeepVA,Vinyals2014ShowAT} is used to produce a caption given the generated image.
In addition to the adversarial image and image-text matching losses, a cross-entropy based text reconstruction loss is used to align the semantics between input caption and re-description.
See \autoref{fig:mirrogan} for an illustration of MirrorGAN.

Inspired by adversarial inference methods \cite{Donahue2017,Dumoulin2017AdversariallyLI}, Lao et al. \cite{Lao2019} proposed to disentangle style (captured via noise vector) and content (described via text embedding) in the latent space in an unsupervised manner.
In their method, an additional encoder takes in real images and infers the two latent variables (style and content), which are subsequently used to generate an image.
A cycle consistency loss term constrains the encoder and decoder to be consistent with one another.
In addition to the adversarial image loss, they also employ a discriminator to distinguish between joint pairs of images and latent codes.

\subsection{Memory Networks}\label{memory}
DM-GAN \cite{Zhu2019e} is an architecture based on dynamic memory networks \cite{Tai2015ImprovedSR,Sukhbaatar2015EndToEndMN,Miller2016KeyValueMN,Glehre2018DynamicNT}.
DM-GAN consists of an initial image generation stage to synthesize a rough $64 \times 64$ pixel image given the sentence embedding.
A memory writing gate takes initial image and word features as input, computes the importance of each word, and finally writes memory slots by combining word and image features.
Then, a key addressing and value reading step is performed in which the relevant memory slots are retrieved by computing a similarity probability between memory slots and image features.
Afterwards, the output memory representation is computed by a weighted summation over value memories according to the similarity probability.
Finally, the gated response dynamically controls the information flow of the output representation to update the image features.
Similar to previously discussed T2I models, DM-GAN employs the unconditional adversarial image and conditional image-text matching losses.
Additionally, the DAMSM loss \cite{Xu2018} and CA loss \cite{Zhang2017} are used.

\subsection{Adapting Unconditional Models}\label{adapt}

\begin{figure}[t]
    \begin{center}
    \includegraphics[scale=0.3]{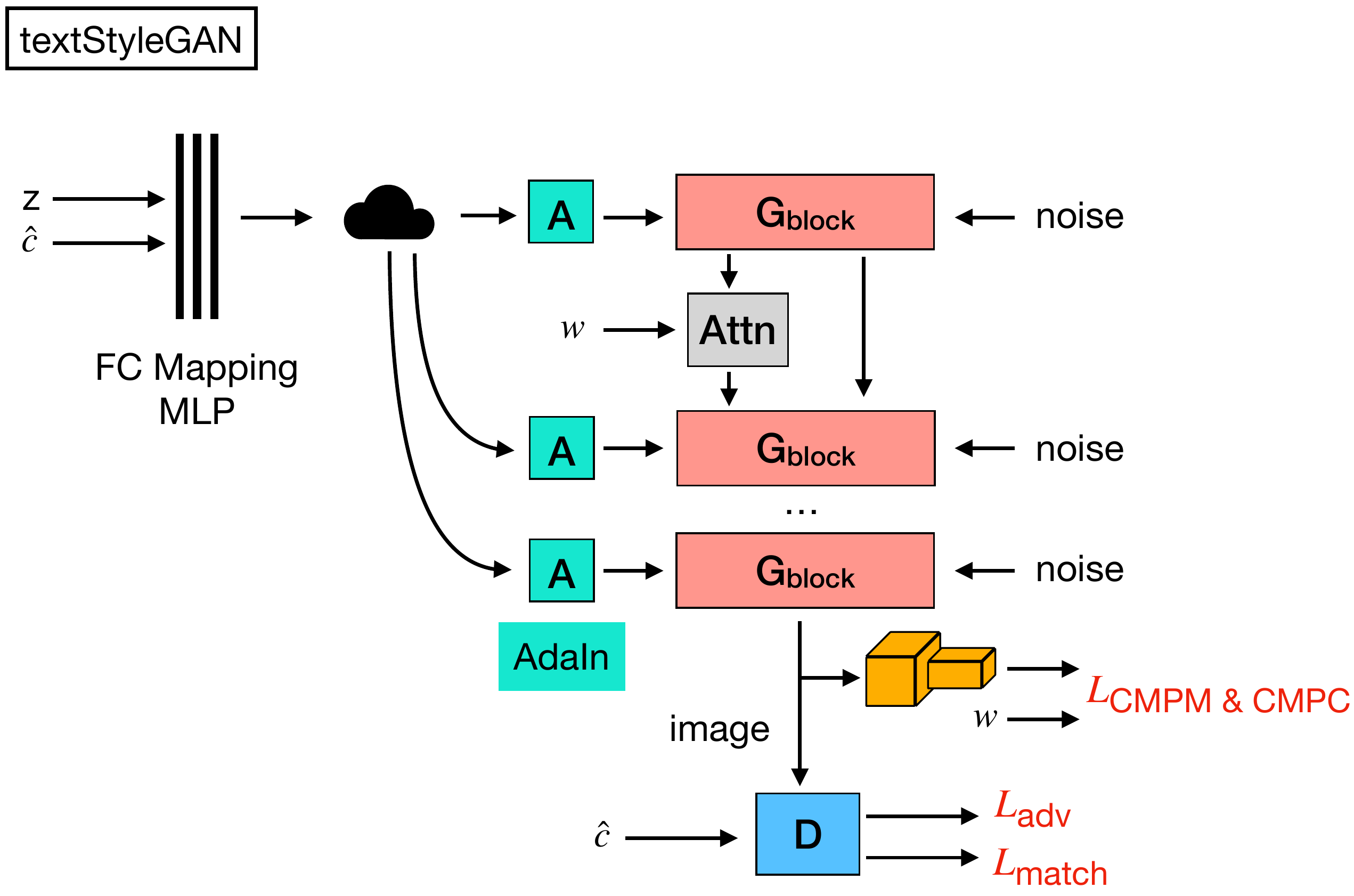}
    \caption{
    Simplified textStyleGAN \cite{textStyleGAN} architecture.
    The noise $z$ and sentence features $\hat{c}$ are first passed through an MLP network to produce an intermediate latent space.
    The intermediate latent vectors $w$ adapt the feature maps via the AdaIn \cite{huang2017arbitrary} operation.
    Given the word features, the attention modules allow the generator to focus on relevant words.
    Each generator block is provided with uncorrelated single-channel noise images for stochastic variation in the generation process.
    The CMPM and CMPC losses \cite{Zhang2018DeepCP} encourage semantic consistency between the generated image and input text.
    See \cite{Karras2018ASG} for more details about StyleGAN.
    }
    \label{fig:textstylegan}
    \end{center}
\end{figure}

Building upon progress in unconditional image generation \cite{Karras2018ASG,BigGAN,ALI}, multiple works proposed to adapt the architecture of these unconditional models for conditional T2I generation.

The authors of textStyleGAN \cite{textStyleGAN} extended StyleGAN \cite{Karras2018ASG}, which can generate images at a higher resolution than other T2I models and allows for semantic manipulation.
The authors proposed to compute text and word embeddings using a pre-trained image-text matching network \cite{Zhang2018DeepCP} similar to the one used in AttnGAN \cite{Xu2018} and concatenate the sentence embedding with the noise vector before performing a linear mapping to produce an intermediate latent space.
Furthermore, they employ attentional guidance using word and image features in the generator.
In addition to unconditional and conditional losses in the discriminator, the cross-modal projection matching (CMPM) and cross-modal projection classification (CMPC) losses \cite{Zhang2018DeepCP} are used to align input captions with generated images.
See \autoref{fig:textstylegan} for an illustration of textStyleGAN.
Image manipulation can be performed by first finding directions in the intermediate latent space corresponding to semantic attributes such as ``smile'' and ``age'' for face images.
Since the intermediate latent space in StyleGAN does not have to support sampling, it has been empirically shown \cite{Karras2018ASG} to unwarp the initial latent code such that the factors of variation become more linear and as a result support semantic image manipulation.

Bridge-GAN \cite{BridgeGAN} employs a progressive growing scheme of the generator and discriminator during training, similar to \cite{Karras2018ProgressiveGO}.
Inspired by \cite{Karras2018ASG}, an intermediate network is used to map the text embedding and noise into a transitional mapping space, and two additional losses based on mutual information are proposed.
The first loss computes the mutual information between the intermediate latent space and input text embedding to guarantee that textual information is present in the transitional space.
The second loss computes the mutual information between generated image and input text to improve the consistency between image and input text.

In \cite{Souza2020EfficientNA}, the authors adapted BigGAN \cite{BigGAN}, an architecture that has previously presented a new state-of-the-art on ImageNet conditioned on class labels, for T2I synthesis.
Furthermore, they proposed a novel Sentence Interpolation method (SI) to create interpolated sentence embeddings using all available captions corresponding to a particular image.
Compared to CA \cite{Zhang2017}, which introduces randomness and optimizes a KL divergence to enforce a Gaussian distribution, SI is a deterministic function.

Similar, TVBi-GAN \cite{TVBiGAN} employed a BiGAN \cite{Donahue2017} architecture by extending the definition of the latent space in ALI \cite{ALI} to project sentence features into it.
Additionally, the authors proposed a gate mechanism inspired by \cite{Zhu2019e} to compute the importance between word features and semantic features before applying attention.
Furthermore, semantics-enhanced batch normalization similar to \cite{Yin2019} is proposed by injecting random noise to stabilize the scale-and-shift operation based on linguistic cues.

In \cite{rombach2020network}, the authors train an invertible network \cite{Dinh2015NICENI,Dinh2017Density} to fuse the pre-trained, expert networks BERT \cite{Devlin2019BERTPO} and BigGAN \cite{BigGAN}, translate between their representations and reuse them for text-to-image synthesis.
This is a very promising research direction to reuse expert networks which are expensive to train for other tasks.

\section{T2I Methods with Additional Supervision}\label{additional_supervision}

In the previous section we discussed T2I approaches that are conditioned on one text description.
However, there are also approaches that incorporate additional supervision.
Models that use more supervision often push the state-of-the-art performance, but they require additional annotations during training.
In the following sections we review methods that use additional inputs such as multiple captions, dialogue data, layout, scene graphs and semantic masks.

\subsection{Multiple Captions}
Since common datasets often contain more than one caption per image, using multiple captions could provide additional information to better describe the whole scene.
C4Synth \cite{Joseph2019} uses multiple captions by employing a cross-caption cycle consistency which ensures that a generated image is consistent with a set of semantically similar sentences.
It operates sequentially by iterating over all captions and improves the image quality by distilling concepts from multiple captions \cite{Joseph2019}.

RiFeGAN \cite{RiFeGAN} treats available images and captions as a knowledge base and uses a caption matching mechanism to retrieve compatible items.
They enrich an input description by extracting features from multiple captions to guide an attentional image generator.
In contrast to \cite{Joseph2019}, RiFeGAN does not need an image captioning network and is executed once instead of multiple times.

\subsection{Dialog}
Motivated by the fact that a single sentence might not be informative enough to describe a scene containing several interacting objects, Sharma et al. proposed ChatPainter \cite{Sharma2018a} to leverage dialog data.
The authors use the Visual Dialog dataset \cite{VisDial}, which consists of 10 question-answer conversation turns per dialogue, and pair it with COCO captions.
The authors experimented with a recurrent and non-recurrent encoder and showed that the recurrent encoder performed better.

Niu et al. \cite{Niu2020ImageSF} proposed VQA-GAN to condition the image generator on locally-related texts by using question answer (QA) pairs from VQA 2.0 \cite{Goyal2017MakingTV}, a dataset built on COCO for visual question answering  (VQA) tasks.
Their method is built upon AttnGAN-OP \cite{Hinz2019GeneratingMO} and consists of three key components: i) a QA encoder that takes QA pairs as input to produce global and local representations, ii) a QA-conditioned GAN that takes the representations from the QA encoder to produce the image in a two-stage process, and iii) an external VQA loss using a VQA model \cite{Benyounes2017MUTANMT} that encourages correlation between the QA pairs and the generated image.
A typical VQA model takes an image and question as input and is trained for classification i.e., to minimize the negative log-likelihood loss to maximize the probability of the correct answer.
Consequently, VQA accuracy can be used as a  metric to evaluate the consistency between input QA pairs and generated images.
Since VQA-GAN is based on \cite{Hinz2019GeneratingMO}, in addition to the QA pairs from VQA 2.0, their model also requires supervision in the form of a layout.

In \cite{frolov2020leveraging}, the authors proposed to leverage VQA data without changing the architecture.
By simply concatenating QA pairs and use them as additional training samples and an external VQA loss, the performance can be improved across both image quality and image-text alignment metrics.
In contrast to \cite{Niu2020ImageSF}, it is a simple, yet effective technique and could be applied to any T2I model.

\subsection{Layout}
There is an increasing interest in the layout-to-image generation task \cite{Layout2Im,LostGAN,LostGANv2,OCGAN} where each object is defined by a bounding box and class label.
It provides more structure to the generator, leads to better localized objects in the image, and has the advantage of allowing user-controlled generation by changing the layout and generated images are automatically annotated.
Naturally, researchers have also tried to combine layout information with text for better T2I.

GAWWN \cite{Reed2016c} conditions on both textual descriptions and object locations to demonstrate the effectiveness of this approach on the CUB-200 Birds dataset.
The follow-up work \cite{Reed2017GeneratingII} extends PixelCNN \cite{Oord2016PixelRN} to generate images from captions with controllable object locations leveraging keypoints and masks.
In \cite{Reed2017ParallelMA}, a parallelized PixelCNN for more efficient inference is used.

In \cite{Hinz2019GeneratingMO}, the location and appearance of objects is explicitly modelled by adding an object pathway to both generator and discriminator.
While the object pathway focuses on generating individual objects at meaningful locations, a global pathway generates a background that fits with the overall image description and layout.
OP-GAN \cite{Hinz2019SemanticOA} extends this by adding additional object pathways at higher layers of the generator and discriminator and also employs an additional bounding box matching loss using matched and mismatched bounding box, image pairs.
OC-GAN \cite{OCGAN} tackles the problem of merged objects and spurious modes by proposing a Scene-Graph Similarity Module (SGSM) similar to DAMSM in AttnGAN \cite{Xu2018}.

\subsection{Semantic Masks}
Another line of research leverages masks to learn the object shapes thereby providing an even better signal to the network.

Hong et al. \cite{Hong2018} obtain the semantic masks in a two-step process: the first step generates a layout from the input description, which is then used to predict object shapes.
It has a single-stage image generator and conditions only on the generated shape and global sentence information.

Obj-GAN \cite{Li2019e} builds upon \cite{Hong2018} and consists of an object-driven attentive generator and an object-wise discriminator.
The generator uses GloVe \cite{Pennington2014GloveGV} embeddings of object class labels to query GloVe embeddings of relevant words in the sentence.
The object-wise discriminator is based on Fast R-CNN \cite{Girshick2015FastR} to provide a signal on whether the synthesized objects are realistic and match the layout and text description.

LeicaGAN \cite{Qiao2019a} has a multiple priors learning phase in which a text-image encoder learns semantic, texture, and color priors, while a text-mask encoder learns shape and layout priors.
These complementary priors are aggregated and used to leverage both local and global features to progressively create the image.
To reduce the domain gap during projection of the input text into an underlying common space, the authors adopted an adversarially trained modality classifier during training.

AGAN-CL \cite{AGANCL} consists of a network which is trained to produce masks, thereby providing fine-grained information such as the number of objects, location, size and shape.
The authors employed a multi-scale loss between real and generated masks, and an additional perceptual loss for global coherence.
In a next step, the image mask is given as input to a cyclic autoencoder, similar to \cite{Zhu_2017_ICCV}, to produce a photo-realistic image.

In \cite{Wang2020ACM}, Wang et al. proposed an end-to-end framework with spatial constraints using semantic layout to guide the image synthesis.
Multi-scale semantic layouts are fused with text semantics and hidden visual features to produce images in a coarse-to-fine way.
At each stage the generator produces an image and additionally a layout to be used by the corresponding discriminator.
The matching aware discriminator from \cite{Reed2016} is extended to also distinguish between matching and mismatching layout-text pairs as well as distinguish real from generated layouts.

Pavllo et al. \cite{Pavllo2020ControllingSA} proposed a weakly-supervised approach by exploiting sparse, instance semantic masks.
In contrast to dense pixel-based masks, sparse instance masks allow easy editing operations such as adding or removing objects because the user does not face the problem of ``filling in wholes''.
Their method is particularly good at controlling fine-grained details of individual objects which is realized by a two-step generation process that decomposes background from foreground.

\subsection{Scene Graphs}
The relationship between multiple objects can often be more explicitly represented by structured text i.e., a scene graph instead of a caption.
For COCO, where scene graph annotations are not provided, a scene graph can be constructed from the object locations using six geometric relationships: ``left of'', ``right of'',``above'', ``below'', ``inside'',  and ``surrounding'' \cite{Johnson2018}.
However, there are also other datasets with more fine-grained scene graph annotations which make this approach very promising (e.g., Visual Genome \cite{VisualGenome} provides on average 21 pairwise relationships per image).

Johnson et al. \cite{Johnson2018} used a graph neural network \cite{Wu2020ACS} to process input scene graphs \cite{Johnson2015ImageRU} and computed a scene layout by predicting bounding boxes and segmentation masks for each object.
The individual object boxes and masks are combined to form a scene layout and subsequently used to produce an image by a cascaded refinement network \cite{Chen2017PhotographicIS}.
Ground-truth bounding boxes and optional masks are used during training, but predicted at test time.

An extension of \cite{Johnson2018} is \cite{ashual2019specifying} which uses segmentation masks.
It separates the layout embedding from the appearance embedding which leads to better control by the user and generated images that better match the input scene graph.
Appearance attributes can either be selected from a predefined set or copied from another image.

In \cite{Vo2019VisualRelationCI}, a scene graph is used to predict initial bounding boxes for objects.
Using the initial bounding boxes, relation units consisting of two bounding boxes are predicted for each individual \textit{subject-predicate-object} relation.
Since each entity could participate in multiple relations, all relation-units are unified and converted into a visual-relation layout using a convolutional LSTM \cite{Shi2015ConvolutionalLN}.
The visual-relation layout reflects the structure (objects and relationships) in the scene graph, and each entity corresponds to one refined bounding box.
Finally, the visual-relation layout is used in a conditional, stacked GAN architecture to render the final image.

PasteGAN \cite{PasteGAN} uses scene graphs and object crops to guide the image generation process.
While the scene graph encodes the spatial arrangements and interactions, the appearance of each object is provided by the given object crops.
Object crops and relationships fused together and then fed into an image decoder to generate the output image.

An interactive framework in \cite{Mittal2019InteractiveIG} extends \cite{Johnson2018} with a recurrent architecture to generate consistent images from an incrementally growing scene graph.
The model updates an image generated from a scene graph by changing the scene graph while keeping the previously generated content as much as possible.
Preserving the previous image is encouraged by replacing the noise passed to the cascaded image generator with the previous image and an additional perceptual loss between the images in the intermediate steps.

\subsection{Mouse Traces}
TRECS \cite{koh2020text} uses mouse traces collected by human annotators in the Localized Narratives \cite{pont2020connecting} dataset which pairs images with detailed natural language descriptions and mouse traces.
The mouse traces provide sparse, fine-grained visual grounding for the descriptions.
Given multiple descriptions and their corresponding mouse traces, TRECS retrieves semantic masks from which the images are generated.

\begin{table}[t]
\small
\begin{center}
    \begin{tabular}{ll}
        \toprule
        Input & Method \\
        \midrule
        caption & \makecell[l]{
        \cite{Reed2016}, \cite{Zhang2017}, \cite{Zhang2019b}, \cite{Zhang2018d}, \\
        \cite{Bodla2018a}, \cite{HfGAN}, \cite{Xu2018}, \cite{Huang2019RealisticIG}, \\
        \cite{Gao2019}, \cite{Tan2019b}, \cite{Li2019f}, \cite{Yin2019}, \\
        \cite{Cha2019}, \cite{PPGN}, \cite{Qiao2019}, \cite{Chen2019d}, \\
        \cite{Lao2019}, \cite{Zhu2019e}, \cite{textStyleGAN}, \cite{BridgeGAN}, \\
        \cite{Souza2020EfficientNA}, \cite{TVBiGAN}, \cite{zhang2021crossmodal}} \\
        
        \rowcolor{gray!10} caption + dialogue & \cite{Sharma2018a}, \cite{Niu2020ImageSF}, \cite{frolov2020leveraging} \\
         
        caption + layout & \cite{Reed2016c}, \cite{Hinz2019GeneratingMO}, \cite{Hinz2019SemanticOA}, \cite{OCGAN} \\
        
        \rowcolor{gray!10} caption + semantic masks & \makecell[l]{\cite{Hong2018}, \cite{Li2019e}, \cite{Qiao2019a}, \cite{AGANCL}, \\
        \cite{Wang2020ACM}, \cite{Pavllo2020ControllingSA}} \\
        
        scene graphs & \makecell[l]{\cite{Johnson2018}, \cite{ashual2019specifying}, \cite{Vo2019VisualRelationCI}, \cite{PasteGAN}, \\
        \cite{Mittal2019InteractiveIG}} \\
        
        \rowcolor{gray!10} multiple captions & \cite{Joseph2019}, \cite{RiFeGAN} \\
        
        multiple captions + mouse traces & \cite{koh2020text} \\
        \bottomrule
    \end{tabular}
    \caption{
    Methods grouped by their supervision. We define ``layout'' as bounding box and class label annotations, and ``masks'' as labelled, instance segmentation masks.
    }
    \label{table:supervision_to_model}
\end{center}
\end{table}

\begin{figure}[tbh]
    \begin{center}
    \includegraphics[scale=0.2]{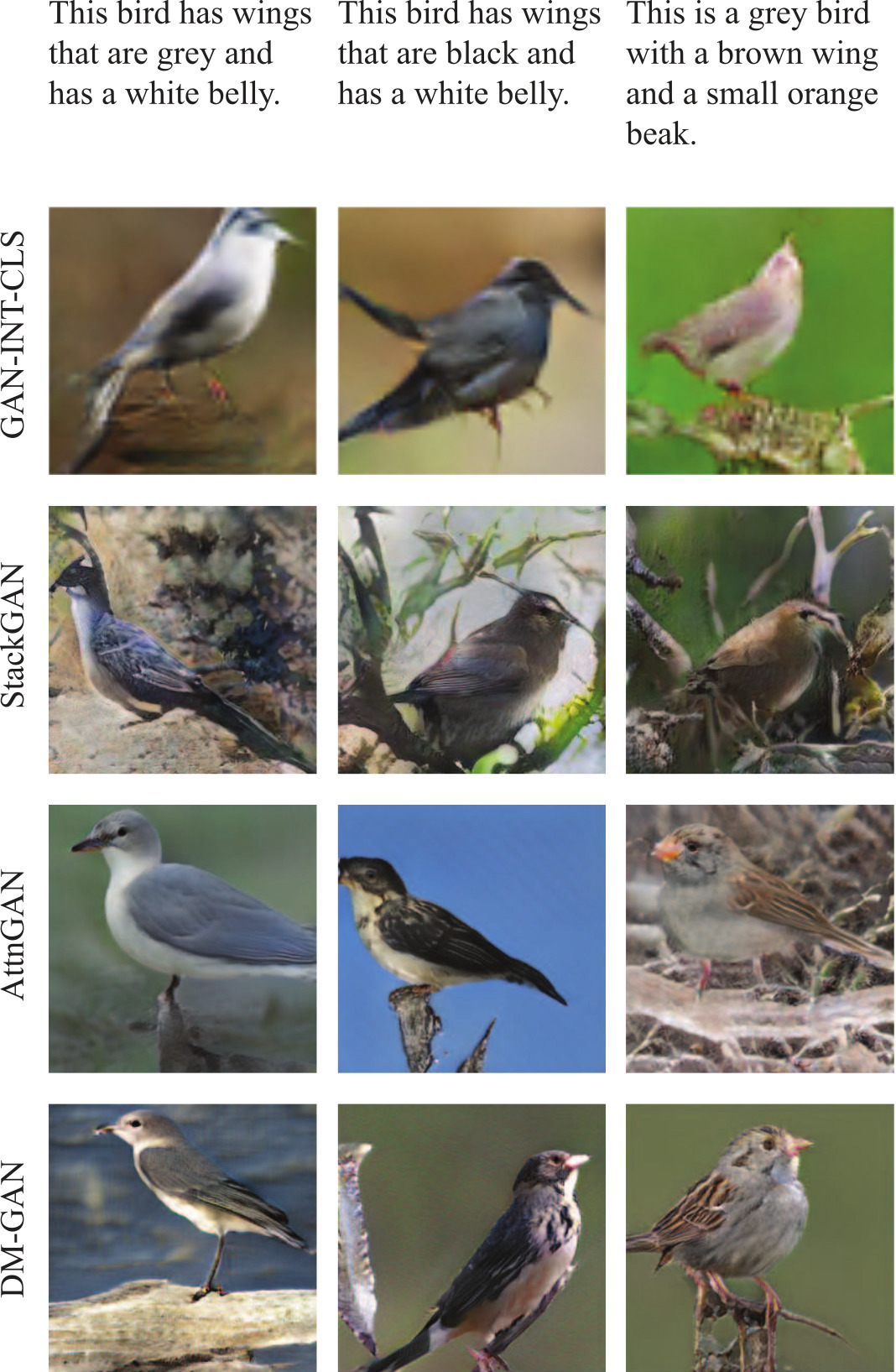}
    \caption{Examples of images generated by models trained on the CUB-200 Birds \cite{Wu2020ACS} dataset. Images are generated by the following models (from top to bottom): GAN-INT-CLS \cite{Reed2016}, Stack-GAN \cite{Zhang2017}, AttnGAN \cite{Xu2018}, and DM-GAN \cite{Zhu2019e}. Figure reproduced from \cite{Zhu2019e}.}
    \label{fig:cub:generated_examples}
    \end{center}
\end{figure}

\begin{figure*}[tbh]
    \begin{center}
    \includegraphics[scale=0.22]{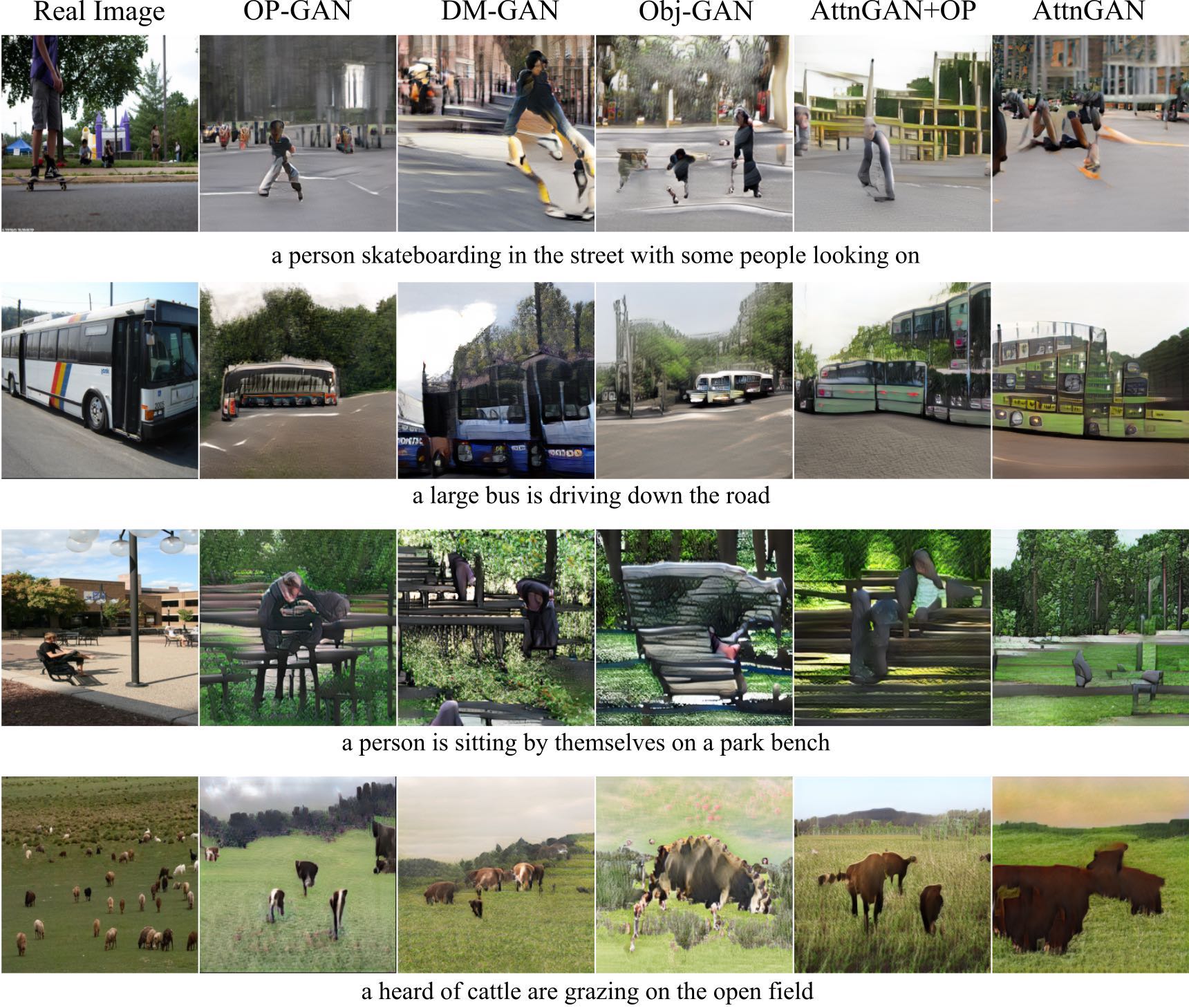}
    \caption{Examples of images generated by models trained on the COCO \cite{COCO} dataset. Images are generated by the following models (from left to right): OP-GAN \cite{Hinz2019SemanticOA}, DM-GAN \cite{Zhu2019e}, Obj-GAN \cite{Li2019e}, AttnGAN-OP \cite{Hinz2019GeneratingMO}, and AttnGAN \cite{Xu2018}. Figure reproduced from \cite{Hinz2019SemanticOA}.}
    \label{fig:coco:generated_examples}
    \end{center}
\end{figure*}

\section{Evaluation of T2I Models}\label{evaluation}

Access to automatic evaluation metrics that correctly assess performance are of utmost importance to gauge improvement and for fair comparison.
Because there are multiple aspects that would resemble a good image (e.g., visual realism and diversity), evaluating generated images is very challenging \cite{Theis2015ANO}.
However, generating realistic images is only one aspect of a good T2I model.
Another important aspect is to assess the semantic alignment between text descriptions and generated images.
In the next sections we revisit which automatic metrics are currently used by the T2I community and how user studies are performed.
Next, we identify and highlight challenges of current evaluation strategies, discuss desiderata of good metrics, and suggest how to evaluate T2I methods with the currently available metrics.
An overview of metrics and what they evaluate is given in \autoref{table:eval}.
We also collect reported results on Oxford-102 Flowers (\autoref{table:oxford_102}), CUB-200 Birds (\autoref{table:cub}), and COCO (\autoref{table:coco}).

\begin{table*}[tbh]
\small
\begin{center}
    \begin{tabular}{lcccccccccc}
        Metric & \rot{Image Quality} & \rot{Image Diversity} & \rot{Object Fidelity} & \rot{Text Relevance} & \rot{Mentioned Objects} & \rot{Numerical Alignment} & \rot{Positional Alignment} & \rot{Paraphrase Robustness} & \rot{Explainable} & \rot{Automatic} \\
        \toprule
        IS \cite{Barratt2018ANO} & \cm & & & & & & & & & \cm \\
        FID \cite{fid} & \cm & \cm & & & & & & & & \cm \\
        SceneFID \cite{OCGAN} & & & \cm & & & & & & & \cm \\
        \midrule
        R-prec. \cite{Xu2018} & & & & \cm & & & & & & \cm \\
        VS \cite{Zhang2018d} & & & & \cm & & & & & & \cm \\
        SOA \cite{Hinz2019SemanticOA} & & & & \cm & \cm & & & & & \cm \\
        Captioning & & & & (\cm) & & & & & & \cm \\
        \midrule
        User Studies & \cm & \cm & \cm & \cm & \cm & \cm & \cm & \cm & \cm & \\
        \bottomrule
    \end{tabular}
    \caption{Overview of commonly used evaluation metrics and desired evaluation aspects. ``Captioning'' refers to metrics used by the image captioning community such as \cite{bleu,meteor,cider}. As they are not visually grounded (just use captions, not also the image to compute the score), we put it in brackets.}
    \label{table:eval}
\end{center}
\end{table*}

\subsection{Image Quality Metrics}
The images generated from textual descriptions should correctly represent the training data distribution.
In the case of commonly used T2I datasets, images should be photo-realistic and diverse.
Many metrics have been proposed to evaluate the image quality of generated images, and we refer to \cite{Borji2018ProsAC} for a detailed review.
In the next paragraphs we revisit and discuss the Inception Score \cite{salimans2016improved} and Fréchet Inception Distance \cite{fid} which are the most frequently used metrics.

    \myparagraph{Inception Score (IS)}
    The IS \cite{salimans2016improved} is computed by classifying generated images with a pre-trained Inception-v3 network \cite{inceptionv3} to get a conditional label distribution $p(y|x)$.
    If the network can produce meaningful images, the conditional label distribution should have low entropy.
    If the network is also able to generate diverse images, the marginal $\int p(y|x=G(z))dz$ should have high entropy.
    In other words, the IS roughly measures how distinctive each image is in terms of classification, and how much variation there is in the generated images overall.
    Both requirements can be measured by computing the Kullback-Leibler (KL) divergence between $p(y|x)$ and $p(y)$, see \autoref{eq:is}.
    The IS is commonly computed from ten splits of a large collection of samples (usually 30k or 50k) and report the average and standard deviation.
    The result is exponentiated to allow for easier comparison:
    \begin{align}\label{eq:is}
    \begin{aligned}
      \text{IS}=\exp(\mathbb{E}_x\,\text{KL}\,(p(y|x)\,||\,p(y))
    \end{aligned}
    \end{align}
    As pointed out in \cite{Hinz2019SemanticOA}, and because of its known weaknesses \cite{Borji2018ProsAC,Barratt2018ANO}, the IS may not be a good measure.
    For example, it can not detect overfitting and can not measure intra-class variation.
    As result, a network that memorizes the training set or only produces one perfect image per class achieves a very high IS.
    Furthermore, it does not take ground truth data into account and uses a classifier pre-trained on the ImageNet dataset, which mostly contains images with one object at the center.
    Hence, it is likely not well suited for more complex datasets where images contain multiple objects such as in COCO.
    
    \myparagraph{Fréchet Inception Distance (FID)}
    The FID \cite{fid} measures the distance between the distribution of real and the distribution of generated images in terms of features extracted by a pre-trained network.
    The FID is more consistent at evaluating GANs than the IS and better captures various kinds of disturbances \cite{fid}.
    Similar to the IS, the FID is usually computed from 30k or 50k of real and generated image samples, using the activations of the last pooling layer of a pre-trained Inception-v3 \cite{inceptionv3} model to obtain visual features.
    To compute the FID, the activations are assumed to follow a multidimensional Gaussian \cite{fid}.
    The FID between real and generated data with mean and covariance of the extracted features $(\mu_r,\Sigma_r)$ and $(\mu_g, \Sigma_g)$, respectively, is then given by \autoref{eq:fid}.
    \begin{align}\label{eq:fid}
    \begin{aligned}
      \text{FID} = \left\|\boldsymbol{\mu}_{r}-\boldsymbol{\mu}_{g}\right\|_{2}^{2} + \operatorname{Tr}\left(\boldsymbol{\Sigma_r}+\boldsymbol{\Sigma}_{g}-2\left(\boldsymbol{\Sigma_r} \boldsymbol{\Sigma}_{g}\right)^{1 / 2}\right)
    \end{aligned}
    \end{align}
    However, the FID assumes that the extracted features follow a Gaussian distribution which is not necessarily the case.
    Furthermore, the estimator of FID has a high bias requiring the same number of samples for fair comparison \cite{mmdgans}.
    The Kernel Inception Distance (KID) introduced in \cite{mmdgans} is an unbiased alternative to the FID, but still has high variance when the number of per-class samples is low \cite{CAS}.
    The FID suffers from the same problem as the IS, in that it relies on a classifier pre-trained on ImageNet.

\subsection{Image-Text Alignment Metrics}
Generating images that look realistic is only one aspect of a good T2I model.
Another important characteristic to assess is whether the generated image aligns with the semantics of the input text description.
The metrics discussed above cannot measure whether the generated image matches the input description.
In the following paragraphs, we review the commonly used R-precision \cite{Xu2018}, Visual-Semantic similarity (VS) \cite{Zhang2018d}, and the recently proposed Semantic Object Accuracy (SOA) \cite{Hinz2019SemanticOA}.

    \myparagraph{R-precision}
    R-prec.\  \cite{Xu2018} measures visual-semantic similarity between text descriptions and generated images by ranking retrieval results between extracted image and text features.
    In addition to the ground truth caption from which an image was generated, additional captions are randomly sampled from the dataset.
    Then, the cosine similarity between image features and the text embedding of each of the captions is calculated and the captions are ranked in decreasing similarity.
    If the ground truth caption from which the image was generated is ranked within the top $r$ captions it is counted as a success.
    In the default setting, R-prec.\ is calculated by setting $r=1$ and randomly sampling 99 additional captions.
    In other words, R-prec.\ evaluates if the generated image is more similar to the ground truth caption than to 99 randomly sampled captions.
    Similar to the previous metrics, R-prec.\ is usually calculated as an average over a large sample (e.g. 30k) of images.
    
    Compared to scores on CUB-200 Birds (\autoref{table:cub}), the R-prec.\ achieved by state-of-the-art models is generally higher for the COCO dataset (\autoref{table:coco}).
    We hypothesize that the images as well as the captions are much more diverse compared to CUB-200 Birds, which makes it easier to distinguish between corresponding and random captions.
    However, R-prec.\ often fails for COCO images, in which a high similarity can be assigned to wrong captions which mention the global background color (e.g., ``snow'' for images with white background) or objects that appear in the center \cite{Hinz2019SemanticOA}.
    
    \myparagraph{Visual-Semantic (VS) Similarity}
    VS similarity proposed in \cite{Zhang2018d} measures the alignment between synthesized images and text by computing the distance between images and text via a trained visual-semantic embedding model.
    Specifically, two mapping functions are learned to map images and text, respectively, into a common representation space.
    Then, the similarity is computed via \autoref{eq:vs}, where $f_t(\cdot)$ is the text encoder, and $f_x(\cdot)$ is the image encoder.
    \begin{align}
    \label{eq:vs}
      \text{VS}=\frac{f_t(t)\cdot f_x(x)}{||f_t(t)||_2 \cdot ||f_x(x)||_2}
    \end{align}
    
    The VS has not been widely adopted by the community and there are only a few reported results.
    A problem of the VS score is that the standard deviation is very high even for real images.
    Therefore, it does not yield a very precise way of evaluating the performance of a model.
    Another challenge that hinders easy comparison arises from using different pre-trained models to compute the VS similarity.
    
    \myparagraph{Captioning Metrics}
    To measure the relevance of generated image and text, \cite{Hong2018,Hinz2019SemanticOA} employ an image caption generator \cite{Vinyals2014ShowAT} to obtain captions for the generated images, and report standard language metrics such as BLEU \cite{bleu}, METEOR \cite{meteor}, and CIDEr \cite{cider}.
    Generated captions should be similar to the input captions that were used to generate the images.
    The hypothesis is that these proxy metrics favour models that produce images which reflect the meaning of the input caption.
    Reported CIDEr scores are shown in \autoref{table:soa}.
    However, it is possible that very different captions correctly describe the same image.
    Furthermore, many of these metrics rely on n-gram overlap and hence may not correlate with human judgement \cite{Hinz2019SemanticOA}.
    
    \myparagraph{Semantic Object Accuracy (SOA)}
    Hinz et al. proposed SOA to evaluate individual objects specifically mentioned in the description within an image using a pre-trained object detector \cite{Hinz2019SemanticOA}.
    For example, we can infer from the caption ``a dog sitting on a couch'' that the image should contain a recognizable dog and a couch, and hence an image detector should be able to detect both objects.
    More specifically, they propose two metrics: SOA-C reports the recall as a class average (i.e., in how many images per class the given object was detected), and SOA-I reports the image average (i.e., in how many images a desired object was detected).
    The authors constructed a list for viable words in the caption for each label and another list containing excluded strings (e.g. ``hot dog'' for ``dog'') in addition to a list of false positive captions.
    In contrast to \cite{Sah2018} which also proposed a detection based evaluation metric, SOA does take the caption into account.
    \autoref{table:soa} shows SOA scores as reported in \cite{Hinz2019SemanticOA}.

    Although SOA is based on words mentioned in the caption, it assumes rather objective and rigid descriptions, where the description is roughly a list of words of visible objects, and hence might not be well suited to evaluate meaning, interaction and relationship between objects, as well as possible subjectivity.
    The authors acknowledged the fact that an image may contain many objects not specifically mentioned in the caption and hence proposed only to focus on false positives, abstaining from calculating a false negative rate \cite{Hinz2019SemanticOA}.
    
\subsection{User Studies}
The metrics presented above are heuristics that have been shown to correlate (to a degree) with human judgements.
To achieve more reliable results, some works additionally perform user studies for verification.
Most studies follow a common structure:
first, each evaluated model generates a certain number of images from a certain number of randomly sampled captions.
Users are then presented with a caption and the generated image(s) from each of the evaluated models.
The users then either have to pick the ``best'' image or rank images from best to worst.
While many user studies are set up in a similar way, there is currently no clear guideline for how these user studies should be structured and evaluated.
Hence, user studies can differ across a number of fundamental factors such as the number of samples, number of users, number of models, specific instructions made to the users, time limitations, and what is finally reported.
Instructions can vary between choosing the ``best'' without specifying what it means, to precise directives such as rating whether objects are identifiable and/or match the input description.
For example, users have been asked to rank images based on the relevance of text \cite{Hong2018}, to select the image which best depicts the caption \cite{Text2Scene,Hinz2019SemanticOA}, to rate whether any one object is identifiable, and how well the image aligns with the text given \cite{Sah2018}, to select the more convincing image and the one which is more semantically consistent with the ground truth \cite{Qiao2019a}.
While some report average ranks, others report the ratio of being ranked first.

As we can see, user studies are not always set up in the same way and, hence, are difficult to compare.
While different setups make comparisons between various user studies unreliable, we highlight that none of the performed studies took real images as an option into account which is just another indication, that current models still struggle to generate images of complex scenes that can fool humans.
Furthermore, performing user studies can be expensive and time consuming.

\subsection{Challenges of Current Techniques}
After introducing the most commonly used evaluation metrics and strategies, we now discuss challenges and shortcomings of these approaches.

    \myparagraph{Higher Scores than Real Images}
    As can be seen in \autoref{table:coco} and \autoref{table:soa}, current models already reach the upper-bound performance in terms of IS, R-precision, and CIDEr given by real images on the COCO dataset.
    This circumstance is misleading given that the generated images are still very unrealistic, and indicates that these metrics might not be reliable.
    The IS can be saturated, even overfitted, and might simply be improved by using a larger batch size \cite{Li2019e}.
    Hinz et al. \cite{Hinz2019SemanticOA} have already observed that the R-prec.\ scores are much higher for some models than for real images and hypothesized that this may be because many of the current models use the same text-encoder during training as well as for final R-prec.\ evaluation.
    Therefore, the models might overfit this metric already during training.
    This problem has also been observed in \cite{zhang2021crossmodal}, and the authors proposed to evaluate the R-prec.\ using a different model which was pre-trained on the large Conceptual Captions \cite{sharma2018conceptual} dataset which is disjoint with COCO.
    In contrast, the reported FID, VS similarity, and SOA scores of current methods are worse than the scores computed on real COCO images, which is in accordance to the approaches still having problems synthesizing individual, sharp objects.
    Due to the high standard deviation in VS results, SOA is likely more meaningful to gauge improvement of future approaches even though it is also just an approximation of human judgement.

    \myparagraph{Single Object Images vs Complex Scenes}
    The IS and FID both use a Inception-v3 network pre-trained on ImageNet, which leads to problems when applied to complex scene images with multiple objects as in the case of COCO.
    In \cite{Hinz2019SemanticOA}, the authors find that the IS has interesting failure cases for images with multiple objects such as assigning the same class to very different images and scenes (bad diversity) or having high entropy in its output layer possibly due to multiple, not centered objects (bad objectiveness).
    One way to mitigate this problem is to apply the IS and FID on object crops.
    In \cite{OCGAN}, the authors train a layout-to-image generator, and proposed SceneFID, which corresponds to the FID applied to object crops as identified by the input bounding boxes.
    This could potentially be adapted even for models that do not take layout as input by using a pre-trained object detector to locate objects.

    \myparagraph{Inconsistent Scores}
    The current literature reports many, often very different, scores for the same model.
    In \autoref{table:reproduced_values}, we collect reported results for multiple models and show that they can vary drastically.
    For example, we found reported FID scores between 35.49 and 28.76 for AttnGAN \cite{Xu2018}, IS scores between 32.43 and 30.49 for DM-GAN \cite{Zhu2019e}, and FID scores between 36.52 and 17.04 and R-prec.\ scores between 91.91\% and 83.00\%, respectively, for Obj-GAN \cite{Li2019e}.
    This suggests that the metrics, even though official implementations exist, are not applied consistently.
    
    While it has been known that the scores can vary depending on the used implementation, image resolution, and number of samples, many inconsistencies are hard to resolve.
    Often occurring problems are that the evaluation procedures are not explained precisely and that the code, if open-sourced, does not contain evaluation code.
    Furthermore, code of baseline methods can be updated and achieve scores different from the ones reported in a paper.
    Most differences are subtle and do not do change the overall ranking, but others are hard to ignore and put the validity and fairness of comparisons into question.
    To improve reproducibility, we encourage researchers to provide precise descriptions of their evaluation procedure, explain possibly existing differences, and to also open-source their evaluation code.

    \myparagraph{Ranking of Models}
    As shown before, user studies are not always set up in the same way which makes comparisons across different studies difficult.
    However, user studies revealed that not all current metrics rank the models as users would.
    This is problematic because we aim to have automatic metrics that correlate with human judgement and allow for accurate and meaningful model rankings.
    For example, in \cite{Text2Scene} the authors observed that users preferred their model when compared against competing methods by a large margin while the IS, and Captioning Metrics showed otherwise.
    Also, the user study in \cite{Hinz2019SemanticOA} showed that FID and SOA matched the user ranking more closely than the IS, R-precision, and CIDEr metrics.

\subsection{Desiderata of Future Metrics}

Developing good automatic metrics is difficult and given the various aspects a generative model could be optimized for, it is very unlikely there will be consensus about the one and only good measure \cite{Borji2018ProsAC}.
Nevertheless, thinking about the desired properties of future metrics can serve as a proxy to compare various metrics with each other and guide future research.
Roughly speaking, a good T2I should be able to both generate high quality images and generate images that align with the input description.
In terms of image quality, we refer the reader to \cite{Borji2018ProsAC} in which the author provides a comprehensive list of desired properties of measures when evaluating generated images such as favouring models with a) high image fidelity and diversity, b) disentangled representations, c) well-defined bounds, d) invariance to small transformations, e) high agreement with human judgement and ranking, and f) low sample and computational complexity.

In terms of image-text alignment, it is difficult to define what precisely it should mean for an image to be aligned with the input description.
Generating images that are ``semantically consistent'', ``fit'', ``match'', or ``correctly reflect'' the input text can be similarly ambiguous expressions.
This is further complicated by the fact that many different captions can correctly describe images depicting complex scenes.
In fact, it might be necessary to first study what exactly makes users prefer one image over an another (especially if both are quite unrealistic).
Despite these difficulties, the following is an attempt to list desired properties which are specifically targeted at evaluating image-text alignment.
Good T2I evaluation should include metrics that:

\begin{itemize}
    \item evaluate whether mentioned objects are depicted and recognizable;
    \item evaluate whether objects are generated according to numerical and positional information in the input description;
    \item evaluate whether the image can correctly be described by the input description;
    \item evaluate whether the model is robust to small changes in the input description (e.g., replacing individual words, or using paraphrases);
    \item are explainable, i.e.\ specify what makes the image not ``aligned'' with the input.
\end{itemize}

As can be seen in \autoref{table:eval}, we currently do not have image-text metrics that evaluate many of the desired aspects.
R-precision, VS similarity, and SOA are only proxies which might not correlate very well with human judgements across the various properties we would like to evaluate.

\subsection{Suggestions to Evaluate T2I Models}
After discussing the current state of evaluation techniques and desiderata of future metrics, it becomes apparent that evaluation is still a very difficult problem and did not necessarily become easier with the proposal of many recent approaches.
In fact, it might have even added to the problem by giving false confidence about the true performance of a method.
While fair and standardized user studies are as of now the only true way to evaluate the performance of a model, we want to suggest how to best use the currently available metrics using our current knowledge about their properties:

\begin{itemize}
    \item we suggest to use the FID to evaluate the visual quality of images and measure the distance to the real data distribution;
    \item we suggest to additionally use the SceneFID on cropped objects if object locations are provided;
    \item we suggest to use SOA (where applicable) and user studies to evaluate the image-text alignment between images and corresponding captions;
    \item we suggest to be precise at describing how the scores were obtained and clearly indicate whether baseline scores were copied from a reference or re-computed;
    \item we suggest to provide a thorough description of how user studies were setup with details about the number of samples/models/users, and specific instructions;
    \item we encourage researchers to open-source not just training, but also evaluation code and report the used implementation and version.
\end{itemize}

\section{Discussion \& Challenges}\label{discussion}
In the last chapters we reviewed state-of-the-art T2I methods, currently used evaluation techniques, desiderata of future metrics as well as how to evaluate T2I models with the currently available ones.
Next, we summarize the current progress in this field, highlight challenges, and discuss future research directions.

\subsection{Model Architecture}\label{discussion_architectures}
Synthesizing images from text has experienced a lot of progress.
Compared to a rather simple architecture in 2016 with one generator, one discriminator and a basic adversarial GAN loss during training, current methods often employ a multi-stage pipeline and several contributing losses.
Starting from generated images having a low resolution, we can now generate realistic looking flower, bird, and high-resolution face images.
While there is also a large improvement on more challenging datasets like COCO, the produced images, and in particular individual objects, lack fine-grained details and sharpness.

Compared to high-quality and high-resolution results currently achieved on single object images, generating complex scenes with multiple objects remains difficult.
The architectural development of T2I methods reflects the general progress made in the field of deep learning (e.g., attention mechanisms, cycle consistency, dynamic memory, Siamese architectures).
More importantly, current T2I approaches have shown successful adaptations of state-of-the-art unconditional image generation models for T2I.
Therefore, building upon progress made in the unconditional image generation domain and investigating better adaptations for conditional image generation might be more efficient than designing special architectures for T2I.

\myparagraph{Importance of Text Embeddings}
One neglected but rather interesting aspect is to investigate the importance and influence of various linguistic aspects in the descriptions such as sensitivity to grammar, positional, and numerical information.
Since the introduction of AttnGAN \cite{Xu2018}, many following works used the same, pre-trained, text encoder to obtain text embeddings, and there has been little investigation into how the embedding quality affects the final T2I performance.
Recent works \cite{wang2020faces,Pavllo2020ControllingSA} leverage transformer-based encoders such as a pre-trained BERT \cite{Devlin2019BERTPO} model to obtain text embeddings for T2I.
In \cite{rombach2020network}, the authors used an invertible network \cite{Dinh2015NICENI,Dinh2017Density} to translate between BERT \cite{Devlin2019BERTPO} and BigGAN \cite{BigGAN} to tackle T2I.
Another interesting direction could be to build upon successes of vision-and-language models \cite{Lu2019ViLBERTPT,Li2019VisualBERTAS,Tan2019LXMERTLC} which have recently shown remarkable progress when fine-tuned on downstream tasks.

\myparagraph{Other Generative Methods}
Current T2I methods are heavily based on GANs which still have many open problems despite the remarkable progress during the last few years \cite{odena2019open}.
Hence, one possible future research direction could be to investigate and build upon progress made with other generative models such as Variational Autoencoders (VAEs) \cite{VAEKingma2013,Razavi2019GeneratingDH}, autoregressive models \cite{Oord2016PixelRN,VanDenOord2016,Menick2019GeneratingHF}, flow-based models \cite{Dinh2015NICENI,Dinh2017Density,Kingma2018GlowGF}, score matching networks \cite{hyvarinen2005estimation,song2019generative,jolicoeur2020adversarial,song2020scorebased}, and transformer-based models \cite{imageTransformer,chen2020generative,esser2020taming,unpublished2021dalle}.
However, comparing different generative models using the IS and FID might be unfair since they penalize non-GAN models \cite{CAS}.
Hence, future evaluation strategies should be model-agnostic to enable reliable comparisons.

\myparagraph{Lack of Scene \& Object Understanding}
Although currently used datasets provide multiple textual descriptions, most often they are semantically very similar (with the notable exception of the COCO dataset).
Moreover, single sentence descriptions are possibly insufficient to describe a complex scene such as in the case of COCO images.
Current models struggle to generate images of multiple, interacting objects and various scenes directly because the captions may not provide enough information.
In fact, current methods seem to fail at modelling simple objects by trying to generate whole scenes because they lack the understanding that scenes are composed of objects.
Approaches such as \cite{Li2019e} and \cite{Hong2018} therefore decompose T2I into text-to-layout-to-image and text-to-mask-to-image, respectively, to guide the generation process.
Another approach taken by \cite{Pavllo2020ControllingSA} is to use instance masks for the desired objects and split the image generation process into foreground (objects), and background synthesis before blending them into the final image (similar to \cite{turkoglu2019layer} and iterative generation as in \cite{Cheng2018SequentialAG,ElNouby2019TellDA,Sharma2018a}).

\subsection{Datasets}\label{discussion_datasets}
Large, high quality datasets are fundamental to the success of deep learning methods.
In the following, we discuss the state of currently used datasets and where future work might enable further advancements of the fields.

\myparagraph{Single Object Datasets}
Many of the recent methods do not report results on the Oxford-102 Flowers dataset anymore.
It is similar to CUB-200 Birds in that images depict a single object only.
However, compared to CUB-200 Birds, there are slightly fewer images, and just 100, as opposed to 200, different object categories.
Hence, using CUB-200 Birds to evaluate T2I methods on single object dataset is enough, and Oxford-102 Flowers does not yield more meaningful insights.
Another approach could be to use the high-resolution human face dataset CelebA-HQ \cite{Karras2018ProgressiveGO} for T2I as was done in \cite{wang2020faces,textStyleGAN}.
Unfortunately, the captions or code to reproduce the captions from the provided attribute labels are not open-sourced as off now.
Since current generative models can synthesize highly realistic images when trained on single object datasets, the focus of evaluation should be on image-text alignment.

\myparagraph{Low Image Resolution of Multiple Object Datasets}
One drawback of currently available datasets of complex scenes depicting multiple objects is the low image resolution.
As of now, we still lack generative methods that can be trained to synthesize photorealistic images of complex scenes with multiple, interacting objects.
Although image quality is currently the bottleneck, it might soon be necessary to collect a high-resolution dataset of images with multiple objects in diverse settings to enable further progress and build practical applications.

\myparagraph{Visually Grounded Captions}
Building upon the idea of locally-related texts \cite{Niu2020ImageSF}, future work might consider allowing to provide textual descriptions for individual regions in the image.
An interesting recent approach considers captions which are paired with mouse traces \cite{koh2020text} from the Localized Narratives \cite{pont2020connecting} dataset, which provide sparse, fine-grained visual grounding for textual descriptions.
Another starting point could be the Visual Genome \cite{VisualGenome} dataset, which contains descriptions of individual image regions.

\myparagraph{Objectivity vs.\ Subjectivity}
One aspect that has not been addressed yet is the incorporation of subjectivity.
A recent study \cite{Blandfort2017ImageCI} analyzes human generated captions and observes that captions that simply describe the obvious image contents are not very common.
This also raises questions regarding a good dataset and requirements for T2I.
Moreover, current datasets are better suited for image captioning, since the captions were collected by asking humans to describe images.
To get insights into how humans interpret textual descriptions and draw mental pictures, one might need to collect images created by humans given a description (similar to how Eitz et al. collected sketches drawn by humans given an object category \cite{Eitz2012HowDH}).

\myparagraph{Limited Cross-Modal Associations}
Another problem stemming from the fact that the T2I community relies on image captioning datasets are the one-sided annotations.
In other words, current datasets provide multiple, matching captions for one image, and such annotations could potentially help to improve T2I models (e.g., via a curriculum learning scheme).
But it is also possible to correctly assign the same caption to describe multiple different images.
This problem is addressed in \cite{parekh2020crisscrossed} by extending the COCO annotations and providing continuous semantic similarity ratings for existing image-text pairs, new pairs, and ratings between captions.
Unfortunately, the Crisscrossed Captions (CxC) \cite{parekh2020crisscrossed} dataset provides ratings only for the COCO evaluation splits.

\myparagraph{Towards Multilingual T2I}
Furthermore, current datasets are limited to the English language.
To increase the practical usefulness of T2I models, future work could consider collecting descriptions of other languages and analyze whether there are differences in how target images are described.
It might even be beneficial for generalization to leverage captions from multiple languages.
A practical T2I should handle input captions from various languages without requiring re-training.

\subsection{Evaluation Metrics}\label{discussion_evaluation}

\myparagraph{Image Quality}
Evaluating the quality, diversity, and semantic alignment of generated images is difficult and still an open problem.
It has become easier with the introduction of IS \cite{salimans2016improved} and FID \cite{fid}, but they have their weaknesses.
Besides the IS and FID, there have been multiple other proposals such the detection based score \cite{Sah2018}, SceneFID \cite{OCGAN}, the classification accuracy score (CAS) \cite{CAS}, precision and recall metrics \cite{Sajjadi2018AssessingGM,Kynknniemi2019ImprovedPA}, and the recently proposed density and coverage metrics \cite{Naeem2020ReliableFA}, which have not yet been adopted by the T2I community.
Similar to \cite{Layout2Im,LostGAN}, we could adopt the LPIPS metric \cite{zhang2018unreasonable} as the Diversity Score (DS) on two sets of generated images from the same captions to specifically evaluate the diversity of generated images.

\myparagraph{Image-Text Alignment}
Images created by a T2I model should also semantically align with the input text.
While current models seem to overfit on the R-precision score, the VS similarity and SOA scores correctly reflect that current models are still far from generating realistic images containing multiple objects.
As of now, we do not have a set of good image-text metrics that provide insights across a number of different aspects.
Therefore, and similar to the image captioning community, a solid evaluation requires a user study.

In terms of future work it might help to join forces with the image captioning community whose goal it is to evaluate the opposite direction: whether the generated caption matches the input image.
In \cite{devries2019evaluation}, a joint Fréchet distance metric is proposed which aims at providing a single score to evaluate various conditional modalities by taking both image and conditioning information into account.
However, the strengths and weaknesses compared to existing text-image metrics are not analyzed, and hence it is unclear whether the approach yields better insights.
Furthermore, current automatic metrics rely on activations extracted from pre-trained models.
Therefore, another promising direction could be to investigate pre-trained, cross-modal vision-and-language models \cite{Lu2019ViLBERTPT,Li2019VisualBERTAS,Tan2019LXMERTLC}.

\myparagraph{Standardized User Studies}
Although the progress on automatic metrics is promising, we currently lack metrics that render user evaluation studies obsolete.
While user studies are sometimes performed, the settings can vary drastically, and they can be time consuming and expensive.
Therefore, a promising future research direction is to standardize user evaluation studies for the T2I community.
Similar to HYPE \cite{HYPE} which standardized user evaluation studies for image quality, the community could benefit from a standardized user evaluation strategy for image-text alignment.

\subsection{Practical Applications}
Image synthesis research is often motivated by practical applications.
Many of these (e.g., image editing and computer-aided design) require fine-grained control (e.g., for interactive and iterative manipulation), and so we believe that future work should also focus on gaining fine-grained control over the image generation process.

\myparagraph{Image Manipulation}
It should be possible to manipulate generated images and edit just some parts of an image without affecting other content.
Recent works by Bau et al. \cite{GanPaint2019,bau2020rewriting} are interesting approaches towards achieving this goal.
On the forefront of image manipulation there are also many works that address text-guided image manipulation \cite{Dong2017,Nam2018,Zhu2020ImageMW,liu2020describe,TextGuidedNeuralImageInpainting,li2020manigan}, which might be a more flexible interface for users than, e.g., editing semantic maps or (a limited amount of) labels.
Since text allows to transfer rich information, future models might need to accumulate and compile an overall representation from multiple, possibly different textual descriptions, similar to how humans draw mental pictures of a scene from both high-level information and fine-grained details.
A study collecting practical requirements (application features users would want) for an optimal T2I model could help the community and give research directions towards practical applications.

\myparagraph{Speech and Video}
Building upon the progress made in T2I, multiple recent works proposed and investigated methods for speech-to-image synthesis (S2I) \cite{Suris_2019_CVPR,Jia2019DirectST,Choi2020FromIT,Wang2020S2IGANSG}.
We believe S2I will receive more attention in the future due to its natural interface which can enable many new interesting and interactive applications.
The S2I community can benefit from the T2I community, since S2I can be realized by replacing the text encoder with a speech encoder and vice versa.
Similarly, generating videos from textual descriptions seems like an obvious future research direction \cite{Li2018VideoGF,Balaji2019ConditionalGW}.
However, evaluating text- and speech-to-video methods comes with it's own challenges, because the individually generated frames should be coherent.

\section{Conclusion}
This review presented an overview of state-of-the-art T2I synthesis methods and commonly used datasets, examined current evaluation techniques, and discussed open challenges.
We categorized existing T2I methods into direct T2I approaches which only use a single textual description as input, and other methods which can use additional information such as multiple captions, dialogue, layout, semantic masks, scene graphs or mouse traces.
While synthesizing images from individual captions has experienced a lot of progress in the recent years, generating images of complex scenes with multiple, possibly interacting objects is still very difficult.
The best image quality is achieved by models which leverage additional information in the form semantic masks, and decompose the generation process into generating foreground objects and background separately, before blending them together.

We also revisited the most commonly used evaluation techniques to assess image quality and image-text alignment.
Evaluating T2I models has become easier with the introduction of automatic metrics such as the IS, FID, R-prec., and SOA.
However, these are only proxies for human judgement and we still require user studies for verification, especially when evaluating image-text alignment and subtle aspects such as numerical and positional information.
Performing user studies comes with its own challenges.
Given that we currently lack a standardized setup, we suggest to provide thorough details of the setup with details about the specific instructions made to the users.

Finally, we offered an in-depth discussion of open challenges across multiple dimensions.
In terms of model architecture, we hope to see more analysis on the importance and quality of text embeddings, the application of other generative models for T2I, and approaches which lead to better scene understanding.
Regarding datasets, we believe that visually grounded captions and dense cross-modal associations could be the keys to learn better representations such as the concept of compositionality.
To enable practical applications of T2I, gaining fine-grained control over the image generation process is important.
Hence, future work should focus on iterative and interactive manipulation and regeneration in addition to synthesis.

Although significant progress has been made, there is still a lot of potential for improvement in terms of generating higher resolution images that better align to the semantics of input text, finding better automatic metrics, standardizing user studies, and enabling more control to build user-friendly interfaces.
We hope this review will help researchers to gain an understanding of the current state-of-the-art and open challenges to further advance the field.

\section*{Acknowledgements}
This work was supported by the BMBF projects DeFuseNN (Grant 01IW17002) and  ExplAINN (Grant 01IS19074), the TU Kaiserslautern PhD program, and the DFG project CML (TRR 169).

\bibliography{bib}

\cleardoublepage

\appendix

\setcounter{table}{0}

\section{Collected Results}
In the following tables we collect results as found in the literature on the three most commonly used datasets.
\autoref{table:oxford_102} contains results on Oxford-102 Flowers, \autoref{table:cub} contains results on CUB-200 Birds, and \autoref{table:coco} contains results on COCO.
\autoref{table:vs} contains contains VS results on all three datasets.
\autoref{table:soa} contains SOA results on COCO.
\autoref{table:reproduced_values} shows that there are multiple, often varying, scores in the literature for the same model.

\begin{table*}[tbh]
\small
\begin{center}
    \begin{tabular}{lrr}
        \toprule
        Model               &  IS $\uparrow$       & FID $\downarrow$           \\
        \midrule
        Real Images         & -                    & -                           \\
        \midrule
        GAN-INT-CLS \cite{Reed2016}     & 2.66 & 79.55  \\
        TAC-GAN \cite{Dash2017}  & 3.45 & -     \\
        StackGAN \cite{Zhang2017}  & 3.20 & 55.28\\
        StackGAN++ \cite{Zhang2019b}     & 3.26 & 48.68  \\
        CVAEGAN \cite{Zhang2018c}   & 4.21 & - \\
        HDGAN \cite{Zhang2018d}      & 3.45 & - \\
        Lao et al. \cite{Lao2019} & - & 37.94 \\
        PPAN \cite{Gao2019}          & 3.52 & - \\
        C4Synth \cite{Joseph2019} & 3.52 & -\\
        HfGAN \cite{HfGAN}   & 3.57 & - \\
        LeicaGAN \cite{Qiao2019a} & 3.92 & - \\
        Text-SeGAN \cite{Cha2019}  & 4.03 & - \\
        RiFeGAN \cite{RiFeGAN}      & 4.53 & - \\
        AGAN-CL \cite{AGANCL}    & \bftab 4.72 & - \\
        Souza et al. \cite{Souza2020EfficientNA} & 3.71 & \bftab 16.47 \\
        \bottomrule
    \end{tabular}
    \caption{
    Results on the Oxford-102 Flowers dataset, as reported in the corresponding reference.
    }
    \label{table:oxford_102}
\end{center}
\end{table*}

\begin{table*}[tbh]
\small
\begin{center}
    \begin{tabular}{lrrr}
        \toprule
        Model         &  IS $\uparrow$ & FID $\downarrow$ & R-Prec. $\uparrow$ \\
        \midrule
        Real Images        & -     & - & - \\
        \midrule
        GAN-INT-CLS \cite{Reed2016} & 2.88  & 68.79 & - \\
        TAC-GAN \cite{Dash2017}    & -    & - & - \\
        GAWWN \cite{Reed2016c} & 3.62 & 67.22 & - \\
        StackGAN \cite{Zhang2017}   & 3.70 & 51.89 & -  \\
        StackGAN++ \cite{Zhang2019b}  & 4.04 & 15.30 & - \\
        CVAEGAN \cite{Zhang2018c} & 4.97 & - & - \\
        HDGAN \cite{Zhang2018d}       & 4.15 & - & - \\
        FusedGAN \cite{Bodla2018a}   & 3.92 & - & - \\ 
        PPAN \cite{Gao2019}           & 4.38 & - & - \\
        HfGAN \cite{HfGAN}  & 4.48 & - & - \\
        LeicaGAN \cite{Qiao2019a} & 4.62 & - & - \\
        AttnGAN \cite{Xu2018}   & 4.36 & - & 67.82 \\
        MirrorGAN \cite{Qiao2019} & 4.56 & - & 57.67 \\
        SEGAN \cite{Tan2019b}   & 4.67 & 18.17 & -\\
        ControlGAN \cite{Li2019f} & 4.58 & - & 69.33 \\
        DM-GAN \cite{Zhu2019e} & 4.75 & 16.09 & 72.31 \\
        DM-GAN \cite{Zhu2019e}\textsuperscript{\textdagger} & 4.71 & 11.91 & 76.58 \\
        SD-GAN \cite{Yin2019}  & 4.67 & - & - \\
        textStyleGAN \cite{textStyleGAN}  & 4.78 & - & 74.72 \\
        AGAN-CL \cite{AGANCL}   & 4.97 & - & 63.87 \\
        TVBi-GAN \cite{TVBiGAN}  & 5.03 & 11.83  & - \\
        Souza et al. \cite{Souza2020EfficientNA}  & 4.23 & \bftab 11.17  & - \\
        RiFeGAN \cite{RiFeGAN}   & \bftab 5.23 & - & - \\
        Wang et al. \cite{Wang2020ACM}      & 5.06 & 12.34 & \bftab 86.50 \\
        Bridge-GAN \cite{BridgeGAN} & 4.74 & - & - \\
        \bottomrule
    \end{tabular}
    \caption{
    Results on the CUB-200 Birds dataset, as reported in the corresponding reference.
    Rows marked with \textsuperscript{\textdagger} indicate updated results in its open-source code.
    }
    \label{table:cub}
\end{center}
\end{table*}

\begin{table*}[tbh]
\small
\begin{center}
    \begin{tabular}{lrrr}
        \toprule
        Model         &  IS $\uparrow$ & FID $\downarrow$ & R-Prec. $\uparrow$        \\
        \midrule
        Real Images \cite{Hinz2019SemanticOA} & 34.88  & 6.09 & 68.58 \\
        \midrule
        GAN-INT-CLS \cite{Reed2016} & 7.88 & 60.62 & -  \\
        StackGAN \cite{Zhang2017} & 8.45 & 74.05 & - \\
        StackGAN \cite{Zhang2017}\textsuperscript{\textdagger} & 10.62 & & - \\ 
        StackGAN++ \cite{Zhang2019b} & 8.30 & 81.59 & - \\
        ChatPainter \cite{Sharma2018a} & 9.74 & - & - \\
        HDGAN \cite{Zhang2018d} & 11.86 & - & - \\
        HfGAN \cite{HfGAN} & 27.53 & - & - \\
        Text2Scene \cite{Text2Scene}  & 24.77 & - & - \\
        AttnGAN \cite{Xu2018} & 25.89 & - & 85.47 \\
        MirrorGAN \cite{Qiao2019} & 26.47 & - & 74.52 \\
        Huang et al. \cite{Huang2019RealisticIG} & 26.92 & 34.52 & 89.69  \\
        AttnGAN+OP \cite{Hinz2019GeneratingMO} & 24.76 & 33.35 & 82.44 \\
        OP-GAN \cite{Hinz2019SemanticOA}& 27.88 & 24.70 & 89.01  \\
        SEGAN \cite{Tan2019b} & 27.86 & 32.28 & - \\
        ControlGAN \cite{Li2019f} & 24.06 & - & 82.43 \\
        DM-GAN \cite{Zhu2019e} & 30.49 & 32.64 & 88.56  \\
        DM-GAN \cite{Zhu2019e}\textsuperscript{\textdagger} & 32.43 & 24.24 & 92.23 \\
        Hong et al. \cite{Hong2018} & 11.46 & - & - \\
        Obj-GAN \cite{Li2019e} & 27.37 & 25.64 & 91.05 \\
        Obj-GAN \cite{Li2019e}\textsuperscript{\textdagger} & 27.32 & 24.70 & 91.91 \\
        SD-GAN \cite{Yin2019} & 35.69 & -  & - \\
        textStyleGAN \cite{textStyleGAN} & 33.00  & - & 87.02 \\
        AGAN-CL \cite{AGANCL} & 29.87 & - & 79.57 \\
        TVBi-GAN \cite{TVBiGAN} & 31.01 & 31.97 & - \\
        RiFeGAN \cite{RiFeGAN} & 31.70 & - & - \\
        Wang et al. \cite{Wang2020ACM} & 29.03  & 16.28 & 82.70 \\
        Bridge-GAN \cite{BridgeGAN} & 16.40 & - & - \\
        Rombach et al. \cite{rombach2020network} & 34.7 & 30.63 & - \\
        CPGAN \cite{liang2020cpgan} & \bftab 52.73 & - & \bftab 93.59 \\
        Pavllo et al. \cite{Pavllo2020ControllingSA} & - & 19.65 & - \\
        XMC-GAN \cite{zhang2021crossmodal} & 30.45 & \bftab 9.33 & - \\
        \bottomrule
    \end{tabular}
    \caption{
    Results on the COCO dataset, as reported in the corresponding reference.
    Rows marked with \textsuperscript{\textdagger} indicate updated results in its open-source code.
    }
    \label{table:coco}
\end{center}
\end{table*}
\begin{table*}[tbh]
\small
\begin{center}
    \begin{tabular}{lrrr}
    \toprule
    Model  & Oxford & CUB & COCO  \\
    \midrule
    Real Images \cite{Zhang2018d}  & 33.6 $\pm$ 13.8 & 30.2 $\pm$ 15.1 & 42.6 $\pm$ 15.7 \\
    \midrule
    GAN-INT-CLS \cite{Reed2016}\textsuperscript{\textdagger} & -                                 & 8.2 $\pm$ 14.7 & - \\
    GAWWN \cite{Reed2016c}\textsuperscript{\textdagger} & -                                 & 11.4 $\pm$ 15.1 & - \\
    StackGAN \cite{Zhang2017}   & 27.8 $\pm$ 13.4 & 22.8 $\pm$ 16.2 & - \\
    HDGAN \cite{Zhang2018d}     & 29.6 $\pm$ 13.1                   & 24.6 $\pm$ 15.7                    & 19.9 $\pm$ 18.3 \\
    HfGAN \cite{HfGAN}          & \bftab 30.3 $\pm$ 13.7            & 25.3 $\pm$ 16.5                    & \bftab 22.7 $\pm$ 14.5 \\
    PPAN \cite{Gao2019}         & 29.7 $\pm$ 13.6                   & 29.0 $\pm$ 14.9                    & - \\
    Bridge-GAN \cite{BridgeGAN}\textsuperscript{\textdagger} & -                                 & \bftab 29.8 $\pm$ 14.6             & - \\
    \midrule
    \midrule
    Real Images \cite{Tan2019b}  & - & 46.3  & 21.2  \\
    \midrule
    AttnGAN \cite{Xu2018}   & - & 22.5 & 7.1 \\
    SEGAN \cite{Tan2019b}   & - & \bftab 30.2          & \bftab 8.9       \\
    \bottomrule
    \end{tabular}
    \caption{
    Reported VS results (higher is better).
    Rows marked with \textsuperscript{\textdagger} indicate results as reported in \cite{BridgeGAN}.
    Results in the second section of the table are from \cite{Tan2019b} computed using a different pre-trained model \cite{Faghri2017VSEIV} for evaluation.
    }
    \label{table:vs}
\end{center}
\end{table*}
\begin{table*}[tbh]
\small
\begin{center}
    \begin{tabular}{lrrr}
        \toprule
        Model                              & SOA-C $\uparrow$       & SOA-I $\uparrow$ & CIDEr $\uparrow$  \\
        \midrule
        Real Images  & 74.97  & 80.84  & 79.5 \\
        \midrule   
        AttnGAN \cite{Xu2018}     & 25.88  & 39.01 & 69.5 \\
        AttnGAN+OP \cite{Hinz2019GeneratingMO} & 25.46 & 40.48 & 68.9 \\
        Obj-GAN \cite{Li2019e} &  27.14  & 41.24 & 78.3 \\
        DM-GAN \cite{Zhu2019e} & 33.44  & 48.03 & \bftab 82.3 \\
        OP-GAN \cite{Hinz2019SemanticOA} & 35.85 & 50.47 & 81.9 \\
        CPGAN \cite{liang2020cpgan} & \bftab 77.02 & \bftab 84.55 & - \\
        XMC-GAN \cite{zhang2021crossmodal} & 50.94 & 71.33 & - \\
        \bottomrule
    \end{tabular}
    \caption{
    Reported SOA-C, SOA-I, and CIDEr results for COCO, as in \cite{Hinz2019SemanticOA}.
    }
    \label{table:soa}
\end{center}
\end{table*}
\begin{table*}[tbh]
\small
\begin{center}
    \begin{tabular}{lrrrr}
        \toprule
        Model & Ref. &  IS $\uparrow$ & FID $\downarrow$ & R-Prec $\uparrow$ \\
        \midrule
        AttnGAN \cite{Xu2018} & \cite{Xu2018} & 25.89 & - & 85.47 \\
                              & \cite{Zhu2019e} & - & 35.49 & - \\
                              & \cite{Tan2019b}  & 25.56 & 34.28 & - \\
                              & \cite{Li2019e} & 23.79 & 28.76 & 82.98 \\
                              & \cite{Hinz2019SemanticOA} & 23.61 & 33.10 & 83.80 \\
                              & \cite{Wang2020ACM} & 23.89 & 28.76 & 82.90 \\
                              & \cite{Qiao2019} & - & - & 72.13 \\
                              & \cite{liang2020cpgan} & - & - & 82.98 \\
                              & \cite{Huang2019RealisticIG} & - & 32.12 & - \\
                              & \cite{frolov2020leveraging} & 26.66 & 27.84 & 83.82 \\
        \midrule
        
         DM-GAN \cite{Zhu2019e} & \cite{Zhu2019e} & 30.49 & 32.64 & 88.56 \\
                                & \cite{Zhu2019e}\textsuperscript{\textdagger} & 32.43 & 24.24 & 92.23 \\
                                & \cite{Hinz2019SemanticOA} & 32.32 & 27.34 & 91.87 \\
                                & \cite{Li2019e} & - & - & 82.70 \\
                                & \cite{liang2020cpgan} & 30.49 & - & 88.56 \\
        \midrule 
        
        Obj-Gan \cite{Li2019e}  & \cite{Li2019e} & 30.29 & 25.64 & 91.05\\ 
                                & \cite{Hinz2019SemanticOA} & 24.09 & 36.52 & 87.84 \\
                                & \cite{Wang2020ACM} & 30.89 & 17.04 & 83.00 \\
                                & \cite{liang2020cpgan} & 30.29 & - & 91.05 \\
        \midrule
        
        OP-GAN \cite{Hinz2019SemanticOA}  & \cite{Hinz2019SemanticOA} & 27.88 & 24.70 & 89.01 \\
                                          & \cite{liang2020cpgan} & 28.57 & - & 87.90 \\
        \bottomrule
    \end{tabular}
    \caption{
    Multiple, often varying, reported results in the literature on the COCO dataset.
    Rows marked with \textsuperscript{\textdagger} indicate updated results in its open-source code.
    }
    \label{table:reproduced_values}
\end{center}
\end{table*}

\end{document}